\documentclass{article}

\usepackage[preprint]{neurips_2026}

\usepackage[utf8]{inputenc} 
\usepackage[T1]{fontenc}    
\usepackage{hyperref}       
\usepackage{url}            
\usepackage{booktabs}       
\usepackage{amsfonts}       
\usepackage{nicefrac}       
\usepackage{microtype}      
\usepackage{xcolor}         
\usepackage{colortbl}       
\definecolor{rowhighlight}{RGB}{225,243,222}      
\definecolor{rowhighlightllama}{RGB}{220,234,250} 
\definecolor{rowhighlightqwen}{RGB}{255,237,214}  

\usepackage{graphicx}
\usepackage{amsmath}
\usepackage{subcaption}
\usepackage{array}
\usepackage{adjustbox}
\usepackage{siunitx}
\usepackage{algorithm}
\usepackage{algorithmic}

\DeclareMathOperator*{\argmax}{arg\,max}

\newcommand{\numw}{0.95cm}
\newcommand{\dsw}{1.90cm}
\newcommand{\dshead}[1]{\parbox[c]{\dsw}{\centering\textbf{#1}}}


\usepackage[most]{tcolorbox}
\tcbuselibrary{skins,breakable}

\newtcolorbox{problembox}{
  enhanced,
  breakable,
  boxrule=0.4pt,
  arc=2mm,          
  left=1.8mm,right=1.8mm,top=1.2mm,bottom=1.2mm,
  colback=white,
  colframe=black
}

\newtcolorbox{promptbox}[1]{%
  enhanced,
  breakable,
  colback=white,
  colframe=black,
  colbacktitle=black,
  coltitle=white,
  fonttitle=\bfseries,
  title=#1,
  arc=2mm,
  boxrule=0.4pt,
  left=2mm,right=2mm,top=1.2mm,bottom=1.2mm,
  before skip=6pt,
  after skip=6pt,
}

\usepackage{cleveref}

\title{Memory-Guided Tree Search with Cross-Branch \\ Knowledge Transfer for LLM Solver Synthesis}

\author{%
  Fatemeh Haji, Javier Delarosa Quiros, Peyman Najafirad \\
  Secure AI and Autonomy Lab \\
  The University of Texas at San Antonio \\
  \texttt{\{fatemeh.haji, javier.delarosaquiros, peyman.najafirad\}@utsa.edu} \\
}

\begin{document}

\maketitle
\begin{abstract}
  Combinatorial optimization (CO) underlies decision-making from logistics to chip design, where infeasible solutions are operationally unusable and small quality gains translate into substantial economic value. Recent work uses large language models (LLMs) to automate \emph{solver synthesis}: generating executable solver programs from natural-language specifications. However, existing tree-search and evolutionary agents refine candidate trajectories in parallel without explicit knowledge transfer, reintroducing the same constraint violations and converging on similar algorithm families. We introduce MEMOIR, a memory-guided tree-search framework with a two-level memory hierarchy: \emph{branch-local memory} preserves execution-grounded refinement details within a branch as it iterates on a single algorithmic design, while \emph{global memory} stores compressed algorithmic and failure-mode summaries across branches. A reflection step at branch termination distills these summaries, enabling cross-branch transfer without polluting future contexts with low-level debugging traces. Across seven CO problems spanning scheduling, routing, packing, and geometric design, MEMOIR achieves 96.7\% solution validity (a 9.2 point gap over the strongest baseline) and improves the average normalized score by 7.3 points at matched per-method execution budget. Over three independent runs on four problems, MEMOIR's run-to-run validity standard deviation is more than an order of magnitude below that of every baseline we evaluated in this setting, suggesting that memory-guided exploration yields consistent improvements rather than reflecting sampling variance.  
\end{abstract}
\section{Introduction}

Combinatorial optimization problems are ubiquitous, spanning domains from logistics and scheduling to chip design and resource allocation. Because these problems are NP-hard, finding optimal solutions is computationally prohibitive \citep{bengio2021machine}, and practitioners have historically relied on hand-crafted, expert-designed heuristics to obtain practical solutions \citep{burke2013hyper}. Two operational realities make automation particularly attractive: in deployment, infeasible solutions are unusable, and even marginal performance improvements translate into substantial economic gains. Yet hand-crafted heuristics are expensive to develop, brittle to problem variants, and difficult to transfer across domains.

A recent line of work addresses these limitations by using LLMs for \emph{automated solver synthesis}, in which generative models produce, from a natural-language problem description or formal program template, a self-contained heuristic program that runs on any instance of the problem class to return a feasible solution \citep{romera2024mathematical, ye2024reevo}. Recent agents explore this program space through iterative sampling and execution-grounded feedback, spanning tree-search methods (AIDE \citep{jiang2025aide}, MCTS-AHD \citep{zheng2025mcts}), evolutionary frameworks (FunSearch \citep{romera2024mathematical}), and reflective evolutionary approaches (ReEvo \citep{ye2024reevo}). These agents maintain multiple candidate trajectories in parallel, but without explicit knowledge transfer across trajectories they repeatedly reintroduce the same constraint violations and revisit equivalent algorithm families across runs, yielding redundant exploration and limited cumulative progress.

Naive transfer is not a fix: reusing raw execution histories pollutes the LLM's context with low-level details and biases generation toward local fixes rather than qualitatively new designs. Existing approaches sit on either side of this tension without resolving it: ReEvo's reflective memory captures refinements within a single algorithm design but does not transfer insights across qualitatively different designs \citep{ye2024reevo}, while AIDE explores a tree of code variants but treats its trajectory journal uniformly, without separating branch-local debugging detail from globally useful algorithmic insight \citep{jiang2025aide}. The two signals must be kept apart: branch-specific refinement traces matter only while a single design is being iterated on, whereas algorithmic and failure-mode lessons should travel across the entire search. We introduce MEMOIR, a memory-guided tree-search framework for LLM-based solver synthesis that realizes this separation through a two-level memory hierarchy and a reflection step that compresses each branch's trajectory before it enters global memory. Tree search is well suited here: each branch corresponds to a single algorithmic design, and branch termination provides a natural point at which to distill branch-local refinements into cross-design lessons. Our contributions are as follows.

{\setlength{\leftmargini}{1.2em}
\begin{itemize}
\setlength{\itemsep}{4pt}\setlength{\parskip}{0pt}
\item \textbf{Two-level memory for solver synthesis.} \emph{Branch-local memory} retains execution-grounded refinements that prevent repeated within-branch failures; \emph{global memory} stores compressed algorithmic and failure-mode summaries that steer subsequent branches toward distinct designs.

\item \textbf{Reflection-based cross-branch knowledge transfer.} A branch-termination reflection distills each trajectory into algorithmic design, failure modes, and avoidance directives, transferring knowledge across branches without low-level debugging traces.

\item \textbf{Reliability under equal budgets.} Across seven classical CO problems at a matched 16-execution budget, MEMOIR reaches 96.7\% validity (+9.2 points over the strongest baseline) and improves normalized score by 7.3 points, with run-to-run standard deviation well below every baseline.
\end{itemize}}

\section{Related Work}
\label{sec:related_work}

\paragraph{Neural combinatorial optimization.} Neural methods train models to directly output solutions for combinatorial optimization problems. Early approaches used RL with policy gradients to learn constructive heuristics \citep{bello2016neural}, later adopting attention mechanisms \citep{kool2019attention} and graph neural networks \citep{khalil2017learning, joshi2019efficient} for routing problems like TSP and VRP. Symmetry-aware training schemes, POMO \citep{pomo} and Sym-NCO \citep{kim2022symnco}, further improved solution quality. However, these learned policies generalize poorly to instance sizes outside their training distribution and become computationally expensive at scale \citep{joshi2022learning}.

\vspace{-0.6em}

\paragraph{Automated modeling.} A parallel line translates natural-language problems into solver-ready formulations consumed by off-the-shelf optimizers, from structured formulation with interactive correction \citep{ramamonjison-etal-2022-augmenting} to modular multi-stage agents \citep{ahmaditeshnizi2024optimus, wang-etal-2025-ormind}, hierarchical tree search \citep{liu2025optitree}, and search-based model synthesis \citep{astorga2025autoformulation}. Formulation correctness remains the bottleneck: small modeling errors invalidate the model and propagate through later stages without clear localization signals \citep{astorga2025autoformulation}.

\vspace{-0.6em}

\paragraph{Evolutionary and tree search for solver synthesis.} A growing line of work synthesizes executable solver code directly from problem specifications, scoring candidates with execution-grounded signals. FunSearch applies evolutionary program search over LLM-generated programs \citep{romera2024mathematical}, EoH co-evolves heuristic descriptions and code with LLMs \citep{liu2024evolution}, ReEvo augments evolution with short- and long-term reflections \citep{ye2024reevo}, and MCTS-AHD performs Monte Carlo Tree Search over algorithm-design states for heuristic synthesis \citep{zheng2025mcts}. More broadly, self-refinement loops improve a single candidate via self-feedback without an explicit search structure \citep{madaan2023self}, tree-search methods structure exploration by branching over intermediate design states \citep{yao2023tree}, AIDE models iterative coding as tree search guided by evaluation \citep{jiang2025aide}, and execution-based verification filters candidates using run-time outcomes \citep{ni2023lever}. Despite differences in search structure, these approaches share a common limitation: each candidate trajectory explores the algorithm space largely in isolation, with selection as the only signal that crosses trajectories. As a result, they repeatedly rediscover failure patterns such as infeasibility bugs or weak algorithmic templates, rather than systematically transferring lessons across iterations.

\vspace{-1.1em}

\paragraph{Memory-augmented agents.} Memory mechanisms have proven valuable in general agent settings. Generative Agents use memory for reflective behavior \citep{park2023generative}, while Voyager maintains a skill library for lifelong learning \citep{wang2024voyager}. Reflexion maintains an episodic buffer of self-generated reflections to improve future decisions \citep{shinn2023reflexion}, and ReEvo distinguishes short-term and long-term reflections within its evolutionary loop \citep{ye2024reevo}. Closer to our hierarchy, Coarse-to-Fine Grounded Memory grounds coarse focus points and fine-grained tips for LLM planning agents \citep{yang-etal-2025-coarse}, but targets interactive planning rather than execution-grounded solver synthesis. However, these approaches do not separate high-level algorithmic design choices that should transfer across iterations from the low-level debugging details that pollute context: Generative Agents and Voyager target general domains, and ReEvo's memory operates within a fixed algorithmic structure.

MEMOIR addresses these challenges through a tree-search--based framework with a two-level memory hierarchy that separates global algorithmic lessons from branch-local implementation details, coupled with reflective compression of design choices and failure modes at branch termination. This enables cross-branch learning without context pollution.

\begin{figure*}[!t]
  \centering
  
  \includegraphics[width=\textwidth,height=0.26\textheight,keepaspectratio]{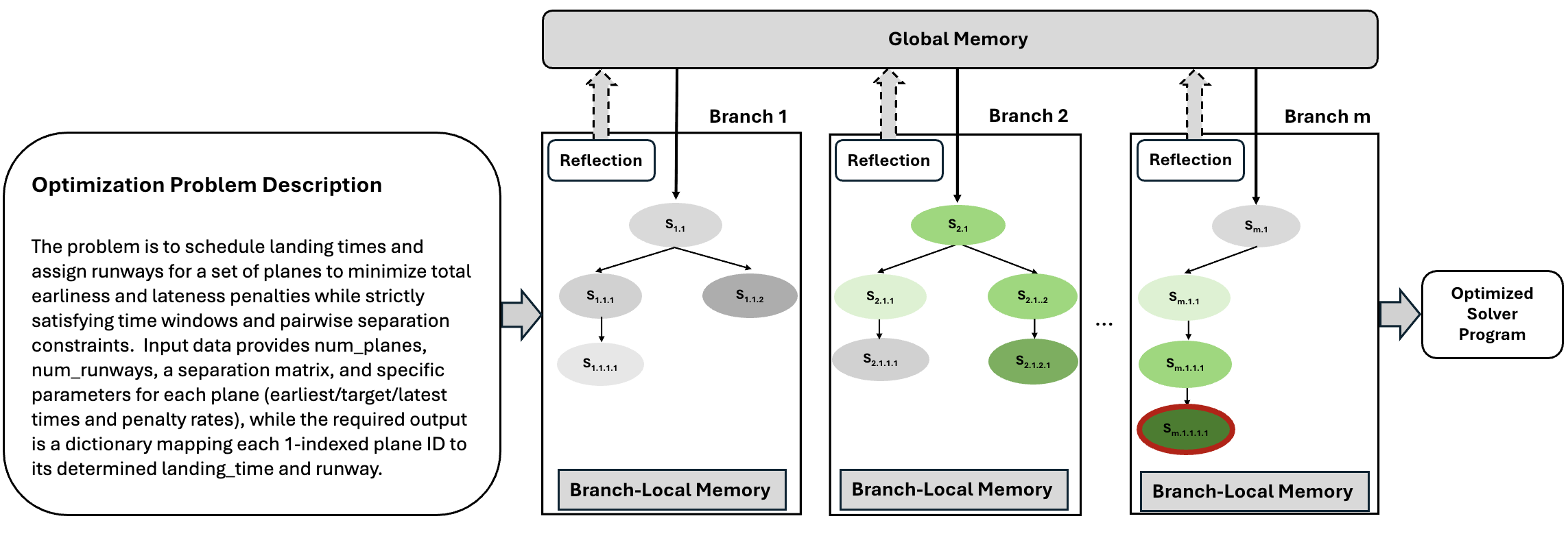}
  \vspace{-6mm} 
  \caption{\textbf{MEMOIR overview.} Each branch starts from a proposed algorithmic design, then runs up to $n$ refinement steps: repair while no valid solver exists in the branch (gray), improve once one does (green). At branch end, reflection compresses the trajectory into a single global-memory entry that steers later branches toward distinct designs. After the execution budget is spent, the highest-scoring valid solver is returned.}
  \vspace{-3mm} 
  \label{fig:main}
  
\end{figure*}

\section{Method}
\vspace{-0.5em}
\subsection{Problem Setting}
\label{sec:problem}

We study automated synthesis of heuristic solvers for combinatorial optimization. A CO problem is specified by a natural-language description $P$, an instance space $\mathcal{X}$, and an instance-level evaluator
$\textsc{Eval}: \mathcal{Y} \times \mathcal{X} \to \mathbb{R}_{\geq 0} \cup \{\bot\}$,
where $\mathcal{Y}$ denotes the solution space and $\bot$ marks infeasibility. We are given a development set $\mathcal{D}_{\text{dev}} \subset \mathcal{X}$ used for synthesis and a held-out test set $\mathcal{D}_{\text{test}} \subset \mathcal{X}$ used only for final reporting.

A \emph{solver} is a Python program $s \in \mathcal{S}$ that, on input instance $x \in \mathcal{X}$ and within a fixed per-instance timeout $T$, generates a stream of candidate solutions $y_1, y_2, \dots \in \mathcal{Y}$ via Python's \texttt{yield} mechanism. We denote by $s(x) \in \mathcal{Y} \cup \{\bot\}$ the last successfully yielded candidate before timeout, with $s(x) = \bot$ if the solver crashes or yields nothing. For brevity, write $h(s, x) = \textsc{Eval}(s(x), x)$, with the convention $h(s,x) = \bot$ whenever $s(x) = \bot$, and let $h^\star(x)$ denote the best-known objective for $x$. For each instance, define the validity indicator and normalized score
\begin{equation}
v(s, x) \;=\; \mathbf{1}\bigl[\, h(s,x) \neq \bot \,\bigr],
\qquad
f(s, x) \;=\;
\begin{cases}
\dfrac{\min\bigl\{|h(s,x)|,\, |h^\star(x)|\bigr\}}{\max\bigl\{|h(s,x)|,\, |h^\star(x)|\bigr\}} & \text{if } v(s,x)=1,\\[10pt]
0 & \text{otherwise.}
\end{cases}
\label{eq:per_instance}
\end{equation}
The symmetric min/max form yields $f(s,x) \in [0,1]$ regardless of whether the underlying problem is minimization or maximization, with $f=1$ when the solver matches the best-known objective and $f \to 0$ as the relative gap grows in either direction; in the degenerate case $|h(s,x)| = |h^\star(x)| = 0$ we set $f = 1$ (exact match).
Per-dataset metrics average these over $\mathcal{D}$: $\textsc{Valid}(s;\mathcal{D}) = |\mathcal{D}|^{-1}\sum_{x \in \mathcal{D}} v(s,x)$ and $\bar f(s;\mathcal{D}) = |\mathcal{D}|^{-1}\sum_{x \in \mathcal{D}} f(s,x)$.\label{eq:dataset_metrics}

\paragraph{Solver synthesis under a query budget}
A solver synthesizer $\mathcal{A}$ is granted a budget of $B$ \emph{executions} on $\mathcal{D}_{\text{dev}}$, where one execution evaluates a single candidate solver on every instance of $\mathcal{D}_{\text{dev}}$. Within this budget, $\mathcal{A}$ iteratively constructs candidate solvers, each conditioned on prior execution feedback, and returns a final selection $\hat s \in \mathcal{S}$. We measure synthesis quality by $\bar f(\hat s;\, \mathcal{D}_{\text{test}})$ and $\textsc{Valid}(\hat s;\, \mathcal{D}_{\text{test}})$.

\subsection{MEMOIR: Memory-Guided Tree Search}
\label{sec:memoir}

MEMOIR formulates solver synthesis as a tree search that separates \emph{exploration} across branches (each instantiating a distinct algorithmic design) from \emph{exploitation} within a branch, refining a single design through targeted edits (\Cref{fig:main}). Given an execution budget $B$ on $\mathcal{D}_{\text{dev}}$, the search exposes two budget knobs: a per-branch depth cap $n$ (exploitation: maximum branch depth, with the initial proposal at depth $1$) and a branch count $m$ (exploration: distinct designs tried).

At any point during search, MEMOIR maintains a memory state $\bigl(\mathcal{M}_{\text{global}},\; \{\mathcal{M}^{(i)}_{\text{local}}\}_{i=1}^{m}\bigr)$.
Each node produced within branch $i$ is a refinement \emph{record}
\begin{equation}
u \;=\; (\pi_u,\, \rho_u,\, v_u,\, f_u,\, E_u),
\label{eq:record}
\end{equation}
where $\pi_u$ is the natural-language algorithmic description of the solver $s_u \in \mathcal{S}$ produced at this node, $\rho_u$ a critic-generated diagnostic comparing $u$ to its parent, $v_u \in \{0, 1\}$ a flag indicating that $s_u$ executed cleanly on \emph{every} instance of $\mathcal{D}_{\text{dev}}$, $f_u = \bar f(s_u;\, \mathcal{D}_{\text{dev}}) \in [0, 1]$ its average normalized score (only the per-instance score in Eq.~\eqref{eq:per_instance} is zeroed on a failed instance, so $f_u$ remains positive whenever some instance executes successfully), and $E_u$ the per-instance execution and evaluation outputs. The solver code $s_u$ itself is retained in a separate artifact store for execution and final return, and is \emph{not} a stored field of branch-local memory. Branch-local memory $\mathcal{M}^{(i)}_{\text{local}}$ accumulates all records produced within branch $i$ and guides repair and improvement of that design; global memory $\mathcal{M}_{\text{global}}$ holds compressed branch-level entries produced by reflection at each branch's termination and acts as a curriculum across designs.

\paragraph{Operators}
Search proceeds via five LLM-implemented operators plus an $\textsc{Execute}$ routine, each conditioned on the relevant memory:

\vspace{-6mm}
\begin{align}
\textsc{Propose}&: P,\, \mathcal{M}_{\text{global}} \;\longrightarrow\; (s_0,\, \pi_0)
&&\text{(branch initialization)}\\[-2pt]
\textsc{Repair}&: P,\, u_{\text{par}},\, \mathcal{M}^{(i)}_{\text{local}} \;\longrightarrow\; (s',\, \pi')
&&\text{(validity recovery)}\\[-2pt]
\textsc{Improve}&: P,\, u_{\text{par}},\, \mathcal{M}^{(i)}_{\text{local}} \;\longrightarrow\; (s',\, \pi')
&&\text{(performance refinement)}\\[-2pt]
\textsc{Critic}&: s,\, s_{\text{par}},\, E,\, E_{\text{par}} \;\longrightarrow\; \rho
&&\text{(parent--child diagnostic)}\\[-2pt]
\textsc{Reflect}&: \mathcal{M}^{(i)}_{\text{local}} \;\longrightarrow\; g_i
&&\text{(branch summary)}
\end{align}
We write $\textsc{Execute}(s; \mathcal{D}_{\text{dev}}) \to (v, f, E)$ for the evaluation that runs $s$ on every instance of $\mathcal{D}_{\text{dev}}$ under timeout $T$ and returns the validity flag, average normalized score, and per-instance outputs. Algorithm~\ref{alg:memoir} (Appendix~\ref{app:algorithm}) composes these operators into the full procedure.

\subsubsection{Branch Initialization: Algorithm Proposer}
\label{sec:proposer}

Each branch begins with a call to the Algorithm Proposer, which jointly generates a high-level algorithmic description $\pi_0$ and a corresponding solver $s_0$ conditioned on $P$ and $\mathcal{M}_{\text{global}}$:
\begin{equation}
(s_0,\, \pi_0) \;\sim\; \textsc{Propose}(P,\, \mathcal{M}_{\text{global}}).
\end{equation}
Conditioning on $\mathcal{M}_{\text{global}}$ steers the proposer away from previously explored designs and their failure modes, encouraging algorithmically distinct branches rather than incremental variants of an earlier candidate. We then execute the initial solver to populate branch $i$'s first record:
\begin{equation}
\begin{aligned}
(v_0, f_0, E_0) &\;\leftarrow\; \textsc{Execute}(s_0;\, \mathcal{D}_{\text{dev}}), \\
\rho_0 &\;\leftarrow\; \textsc{Critic}(s_0,\, \emptyset,\, E_0,\, \emptyset), \\
u_0 &\;=\; (\pi_0, \rho_0, v_0, f_0, E_0), \qquad \mathcal{M}^{(i)}_{\text{local}} \;\leftarrow\; \{u_0\}.
\end{aligned}
\end{equation}

\subsubsection{Within-Branch Refinement: Repair and Improve}
\label{sec:refinement}

After initialization, branch $i$ proceeds with successive refinement steps while further within-design improvement remains plausible from the branch-local memory, up to a maximum branch depth of $n$. At each step, MEMOIR selects a parent record $u_{\text{par}} \in \mathcal{M}^{(i)}_{\text{local}}$ and applies one of two operators based on the branch's current validity:
\begin{equation}
(s',\, \pi') \;=\;
\begin{cases}
\textsc{Repair}\bigl(P,\, u_{\text{par}},\, \mathcal{M}^{(i)}_{\text{local}}\bigr) & \text{if } v_u = 0 \;\;\forall u \in \mathcal{M}^{(i)}_{\text{local}}, \\[3pt]
\textsc{Improve}\bigl(P,\, u_{\text{par}},\, \mathcal{M}^{(i)}_{\text{local}}\bigr) & \text{otherwise.}
\end{cases}
\end{equation}
In the repair case, $u_{\text{par}}$ is sampled from invalid records in proportion to $f_u$; in the improve case, $u_{\text{par}} = \argmax_{u \in \mathcal{M}^{(i)}_{\text{local}},\, v_u = 1} f_u$ is the highest-scoring valid record.
The new solver is then executed and critiqued:
\begin{equation}
\begin{aligned}
(v', f', E') &\;\leftarrow\; \textsc{Execute}(s';\, \mathcal{D}_{\text{dev}}), \\
\rho' &\;\leftarrow\; \textsc{Critic}(s', s_{u_{\text{par}}},\, E',\, E_{u_{\text{par}}}), \\
u' &\;=\; (\pi', \rho', v', f', E'), \qquad \mathcal{M}^{(i)}_{\text{local}} \;\leftarrow\; \mathcal{M}^{(i)}_{\text{local}} \cup \{u'\}.
\end{aligned}
\end{equation}

\paragraph{Repair operator}
$\textsc{Repair}$ receives an invalid parent record, its critic diagnostic, and the full branch-local history; the history lets the operator avoid edits that previously failed to restore validity. Among invalid records, MEMOIR samples the parent in proportion to $f_u$, biasing repair toward solvers that came closest to feasibility; when all invalid records have $f_u = 0$ (no instance executed successfully) we fall back to uniform sampling.
\vspace{-4mm}
\paragraph{Improve operator}
$\textsc{Improve}$ activates once any valid solver exists. Each invocation applies a single focused modification intended to increase $\bar f$ while preserving $v = 1$. Branch-local memory conditions the operator to avoid edits that prior records show reduced $f$ or invalidated previously valid solvers. This single-edit discipline yields incremental, attributable improvements within one algorithmic design and makes regressions easy to localize.

\subsubsection{Cross-Branch Knowledge Transfer: Reflection}
\label{sec:reflection}

Once branch $i$ terminates, its trajectory is compressed into a single global-memory entry via the reflection operator


\begin{equation}
g_i \;=\; \textsc{Reflect}\bigl(\mathcal{M}^{(i)}_{\text{local}}\bigr),
\qquad
\mathcal{M}_{\text{global}} \;\leftarrow\; \mathcal{M}_{\text{global}} \cup \{g_i\}.
\end{equation}
Each entry $g_i$ contains three fields: (i) the branch's algorithmic design and core heuristics, (ii) observed failure or stagnation modes that limited improvement, and (iii) explicit avoidance directives stated as negative constraints for subsequent proposers. All three fields are derived from the branch's execution trace and critic diagnostics rather than from a free-form LLM read of the code, so the entry summarizes observed behavior rather than asserted properties of the program. By design, $g_i$ contains \emph{no} per-instance traces, intermediate code, or low-level debugging artifacts, which is what allows global memory to grow across branches without polluting the proposer's context. Empirically, entries average $\sim$272 tokens across our seven domains, so $|\mathcal{M}_{\text{global}}|$ scales as $\mathcal{O}(m)$ with a small constant and remains bounded under larger branch budgets.

\subsubsection{Final Selection}
\label{sec:final}

After the search terminates, MEMOIR selects the highest-scoring valid record on $\mathcal{D}_{\text{dev}}$, $u^\star = \argmax_{u \in \mathcal{U},\, v_u = 1} f_u$ with $\mathcal{U} = \bigcup_i \mathcal{M}^{(i)}_{\text{local}}$, and returns its associated solver $\hat s$ from the artifact store. If no fully valid record exists, we fall back to $u^\star = \argmax_{u \in \mathcal{U}} f_u$, returning the partially valid solver with the highest average score. We then evaluate $\hat s$ once on $\mathcal{D}_{\text{test}}$ under the same timeout $T$, reporting $\bar f(\hat s; \mathcal{D}_{\text{test}})$ and $\textsc{Valid}(\hat s; \mathcal{D}_{\text{test}})$.

\section{Experiments}
\vspace{-0.5em}
\subsection{Experimental Setup}

We describe the problem domains, baselines, evaluation protocol, and the matched search budget shared by all methods.

\subsubsection{Problem Domains}

We evaluate MEMOIR on seven combinatorial optimization problems chosen to stress two axes: \emph{hard-constraint pressure}, where infeasibility dominates if constraints are mishandled, and \emph{tightly coupled combinatorial structure}, where decisions interact across subproblems. The problems are Aircraft Landing \citep{aircraft_landing_1}, Periodic Vehicle Routing \citep{period_routing}, Container Loading \citep{container_loading}, Container Loading with Weight Restrictions \citep{container_loading_weight}, Resource Constrained Shortest Path \citep{resource_constrained_shortest_path}, Crew Scheduling \citep{crew_scheduling}, and Euclidean Steiner \citep{steiner_problem}. We adopt CO-Bench's natural-language specifications, instance-level evaluators (returning $h(s, x)$, normalized via Eq.~\eqref{eq:per_instance} against the best-known $h^\star$), and dev/test splits ($\mathcal{D}_{\text{dev}}, \mathcal{D}_{\text{test}}$) for comparability with prior solver-synthesis work \citep{sun2026co}; full specifications appear in Appendix~\ref{app:datasets}.

\subsubsection{Baselines}

We compare MEMOIR against three categories of baselines, with LLM-agent methods run under the matched execution budget specified in Section~\ref{sec:hyperparams}.

\noindent\textbf{Classical Solver.} Published scores from expert-designed solvers for each problem, serving as a non-LLM reference.

\noindent\textbf{Direct LLM solver synthesis.} Base LLMs (Llama-3.3-70B-Instruct, Qwen2.5-Coder-32B, GPT-5 Chat) and reasoning LLMs (o3-mini-high, DeepSeek-R1-Distill-Llama-70B, QwQ-32B, GPT-5-mini) are prompted to produce a solver in a single pass, with no iterative search.

\noindent\textbf{LLM-agent methods.} \emph{GreedyRefine} \citep{sun2026co} iteratively rewrites the current best solver, replacing it on improvement; \emph{AIDE} \citep{jiang2025aide} performs tree search over prior attempts, selecting a parent by score and editing it with a single flat trajectory journal as memory; \emph{FunSearch} \citep{romera2024mathematical} runs an islands evolutionary loop, sampling new solvers from top-scoring members of each island; \emph{ReEvo} \citep{ye2024reevo} augments evolutionary search with reflective memory; and \emph{MCTS-AHD} \citep{zheng2025mcts} performs Monte Carlo Tree Search over algorithm-design states.

\subsubsection{Evaluation Protocol}
\label{sec:eval_protocol}

We report two metrics defined in Eqs.~\eqref{eq:per_instance}--\eqref{eq:dataset_metrics}: \textit{Avg}, the mean per-instance normalized score $\bar f(s; \mathcal{D}) \in [0,1]$, with score $0$ assigned to program errors or infeasible solutions; and \textit{Valid}, the fraction of instances where the solver executes successfully and returns a feasible solution. 

\subsubsection{Implementation and Search Budget}
\label{sec:hyperparams}

\paragraph{Budget and per-method search.}
All methods share an execution budget of $B=16$ on $\mathcal{D}_{\text{dev}}$, a compute-limited setting that isolates each method's search efficiency rather than its ability to draw many candidate solvers and pick the best. At evaluation, the final synthesized solver is allotted a per-instance timeout of $T=10$\,s to return a solution for a single test instance, matching the low-latency inference regime relevant to test-time deployment. Hyperparameters are fixed across domains (no per-domain tuning), and solvers execute in a sandboxed Linux environment with Python 3.10.18 (CPU only). For MEMOIR, the per-branch depth cap is $n=5$, but branches terminate early when further improvement is unlikely; we keep opening new branches with the remaining budget, and in our setup branches typically use most of the cap, so $B=16$ yields about $m=4$ branches per run. Per-run token counts and dollar costs are reported in Appendix~\ref{app:cost}; synthesis stays under \$1 per run for every method.
\vspace{-0.6em}
\paragraph{Models and baseline configuration.}
Table~\ref{tab:main_results} reports four MEMOIR variants: \emph{MEMOIR (GPT-5-mini w/ GPT-5 critic)}, which uses \texttt{gpt-5-mini} for Propose, Repair, and Improve and \texttt{gpt-5} for Critic and Reflect, since diagnostic and reflection quality bound the quality of memory entries while solver generation is held to \texttt{gpt-5-mini} for parity with all baselines; and three single-backbone variants \emph{MEMOIR (GPT-5-mini)}, \emph{MEMOIR (Llama-3.3-70B-Instruct)}, and \emph{MEMOIR (Qwen2.5-Coder-32B)}, which run all five operators on the named backbone. For baselines, GreedyRefine, ReEvo, and MCTS-AHD use \texttt{gpt-5-mini}, and FunSearch uses \texttt{gpt-5-mini} with 3 islands. AIDE \citep{jiang2025aide} uses \texttt{gpt-5-mini} for code drafting/modification and \texttt{gpt-5} for execution analysis, matched to MEMOIR's Critic and Reflect so any gap reflects memory design, not feedback quality. Prompts for all MEMOIR operators appear in Appendix~\ref{app:prompts}.

\begin{table*}[!t]
  \caption{\textbf{Per-domain test-set performance across the seven CO problems}
  under a shared execution budget $B{=}16$. Average normalized score (Avg, higher is
  better) and validity rate (Valid, fraction of instances with feasible solutions);
  bold marks the best Overall within each method category.}
  \label{tab:main_results}
  \centering
  \begin{adjustbox}{max width=\textwidth}
  \normalsize
  \setlength{\tabcolsep}{3pt}
  \renewcommand{\arraystretch}{1.30}

  \begin{tabular}{l
    S[table-format=1.4,table-column-width=\numw] S[table-format=1.4,table-column-width=\numw]
    S[table-format=1.4,table-column-width=\numw] S[table-format=1.4,table-column-width=\numw]
    S[table-format=1.4,table-column-width=\numw] S[table-format=1.4,table-column-width=\numw]
    S[table-format=1.4,table-column-width=\numw] S[table-format=1.4,table-column-width=\numw]
    S[table-format=1.4,table-column-width=\numw] S[table-format=1.4,table-column-width=\numw]
    S[table-format=1.4,table-column-width=\numw] S[table-format=1.4,table-column-width=\numw]
    S[table-format=1.4,table-column-width=\numw] S[table-format=1.4,table-column-width=\numw]
    S[table-format=1.4,table-column-width=\numw] S[table-format=1.4,table-column-width=\numw]
  }
    \toprule
    \textbf{Dataset} &
    \multicolumn{2}{c}{\dshead{Aircraft\\Landing}} &
    \multicolumn{2}{c}{\dshead{Periodic Vehicle Routing}} &
    \multicolumn{2}{c}{\dshead{Container\\Loading}} &
    \multicolumn{2}{c}{\dshead{Container Loading\\with Weight Restrictions}} &
    \multicolumn{2}{c}{\dshead{Resource Constrained\\Shortest Path}} &
    \multicolumn{2}{c}{\dshead{Crew\\Scheduling}} &
    \multicolumn{2}{c}{\dshead{Euclidean\\Steiner}} &
    \multicolumn{2}{c}{\dshead{Overall}} \\
    \midrule
    \textbf{Method}
    & \textbf{Avg} & \textbf{Valid}
    & \textbf{Avg} & \textbf{Valid}
    & \textbf{Avg} & \textbf{Valid}
    & \textbf{Avg} & \textbf{Valid}
    & \textbf{Avg} & \textbf{Valid}
    & \textbf{Avg} & \textbf{Valid}
    & \textbf{Avg} & \textbf{Valid}
    & \textbf{Avg} & \textbf{Valid} \\
    \cmidrule(lr){2-3}\cmidrule(lr){4-5}\cmidrule(lr){6-7}\cmidrule(lr){8-9}%
    \cmidrule(lr){10-11}\cmidrule(lr){12-13}\cmidrule(lr){14-15}\cmidrule(lr){16-17}

    Classical Solver
      & .5985 & \multicolumn{1}{c}{--}
      & .1244 & \multicolumn{1}{c}{--}
      & .0970 & \multicolumn{1}{c}{--}
      & .0092 & \multicolumn{1}{c}{--}
      & .7509 & \multicolumn{1}{c}{--}
      & .4550 & \multicolumn{1}{c}{--}
      & .9780 & \multicolumn{1}{c}{--}
      & .4304 & \multicolumn{1}{c}{--} \\

    \addlinespace[2pt]
    \midrule
    \multicolumn{17}{l}{\textbf{Base LLMs}} \\
    \rowcolor{rowhighlightllama}
    Llama-3.3-70B-Instruct
      & .0002 & .1379
      & .0000 & .0000
      & .0000 & .0000
      & .0000 & .0000
      & .7508 & .9167
      & .0000 & .0000
      & .0312 & .2969
      & .1117 & .1931 \\
    \rowcolor{rowhighlightqwen}
    Qwen2.5-Coder-32B
      & .0000 & .0000
      & .0000 & .0000
      & .0000 & .0000
      & .0000 & .0000
      & .7508 & .9167
      & .0000 & .0000
      & .0000 & .0000
      & .1073 & .1310 \\
    \rowcolor{rowhighlight}
    GPT-5 Chat
      & .0345 & .0348
      & .0000 & .0000
      & .0000 & .0000
      & .0000 & .0000
      & .7508 & .9167
      & .5165 & .7586
      & .0000 & .0000
      & .1860 & .2443 \\

    \addlinespace[2pt]
    \midrule
    \multicolumn{17}{l}{\textbf{Reasoning LLMs}} \\
    o3-mini-high
      & .8093 & .9655
      & .0000 & .0000
      & .6542 & 1.0000
      & .0274 & .1057
      & .7508 & .9167
      & .4412 & .6552
      & .4541 & .9922
      & .4481 & .6622 \\
    \rowcolor{rowhighlightllama}
    DeepSeek-R1-Distill-Llama-70B
      & .1670 & .9310
      & .0000 & .0000
      & .0000 & .0000
      & .0000 & .0000
      & .7508 & .9167
      & .0000 & .0000
      & .0156 & .9922
      & .1333 & .4057 \\
    \rowcolor{rowhighlightqwen}
    QwQ-32B
      & .1249 & .9310
      & .0000 & .0000
      & .0441 & .0550
      & .1026 & .4171
      & .7039 & .7500
      & .0000 & .0000
      & .2795 & 1.0000
      & .1793 & .4504 \\
    \rowcolor{rowhighlight}
    GPT-5-mini
      & .6190 & .9310
      & .1350 & .2000
      & .2888 & .9791
      & .0170 & .0743
      & .7508 & .9167
      & .5735 & .6552
      & .3038 & .9922
      & .3840 & .6784 \\

    \addlinespace[2pt]
    \midrule
    \multicolumn{17}{l}{\textbf{LLM Agents}} \\
    GreedyRefine
      & .8035 & .9655
      & .3780 & .4000
      & .7253 & 1.0000
      & .1015 & .4114
      & .7039 & .7500
      & .5545 & .6207
      & .7167 & .9922
      & .5691 & .7343 \\
    AIDE
      & .8228 & .9310
      & .1058 & .2000
      & .7748 & 1.0000
      & .0975 & .6457
      & .7508 & .9167
      & .5600 & .7241
      & .6337 & .9922
      & .5351 & .7728 \\
    MCTS-AHD
      & .8187 & .9655
      & .2990 & .4000
      & .6205 & 1.0000
      & .0161 & .4857
      & .7508 & .9167
      & .6353 & .8621
      & .5917 & .9922
      & .5332 & .8032 \\
    ReEvo
      & .8063 & .9310
      & .3130 & .4000
      & .8338 & 1.0000
      & .1355 & .6543
      & .7508 & .9167
      & .5274 & .6552
      & .6886 & .9922
      & .5793 & .7928 \\
    FunSearch
      & .8788 & .9655
      & .7178 & .8000
      & .8141 & 1.0000
      & .2961 & 1.0000
      & .7039 & .7500
      & .5564 & .6207
      & .6318 & .9922
      & .6570 & .8755 \\
    \rowcolor{rowhighlightllama}
    MEMOIR (Llama-3.3-70B-Instruct)
      & .1132 & .2258
      & .1468 & .4000
      & .0119 & 1.0000
      & .0000 & .0000
      & .7508 & .9167
      & .0000 & .0000
      & .0543 & .9922
      & \bfseries .1539 & \bfseries .5050 \\
    \rowcolor{rowhighlightqwen}
    MEMOIR (Qwen2.5-Coder-32B)
      & .0071 & .2727
      & .0000 & .0000
      & .0162 & 1.0000
      & .0058 & .5400
      & .8447 & 1.0000
      & .3155 & .5500
      & .3124 & .7544
      & \bfseries .2145 & \bfseries .5882 \\
    MEMOIR (GPT-5-mini)
      & .8751 & .9655
      & .7642 & .8000
      & .8084 & 1.0000
      & .3180 & 1.0000
      & .7508 & .9167
      & .5783 & .7586
      & .7243 & 1.0000
      & .6884 & .9201 \\
    \rowcolor{rowhighlight}
    MEMOIR (GPT-5-mini w/ GPT-5 critic)
      & .8668 & .9655
      & .8690 & 1.0000
      & .8364 & 1.0000
      & .3453 & 1.0000
      & .7508 & .9167
      & .6240 & .8966
      & .8183 & .9922
      & \bfseries .7301 & \bfseries .9673 \\
    \bottomrule
  \end{tabular}

  \end{adjustbox}
  \vspace{-5mm}
\end{table*}

\subsection{Main Results}
\label{sec:main_results}

\begin{figure*}[t]
  \centering
  \includegraphics[width=\textwidth]{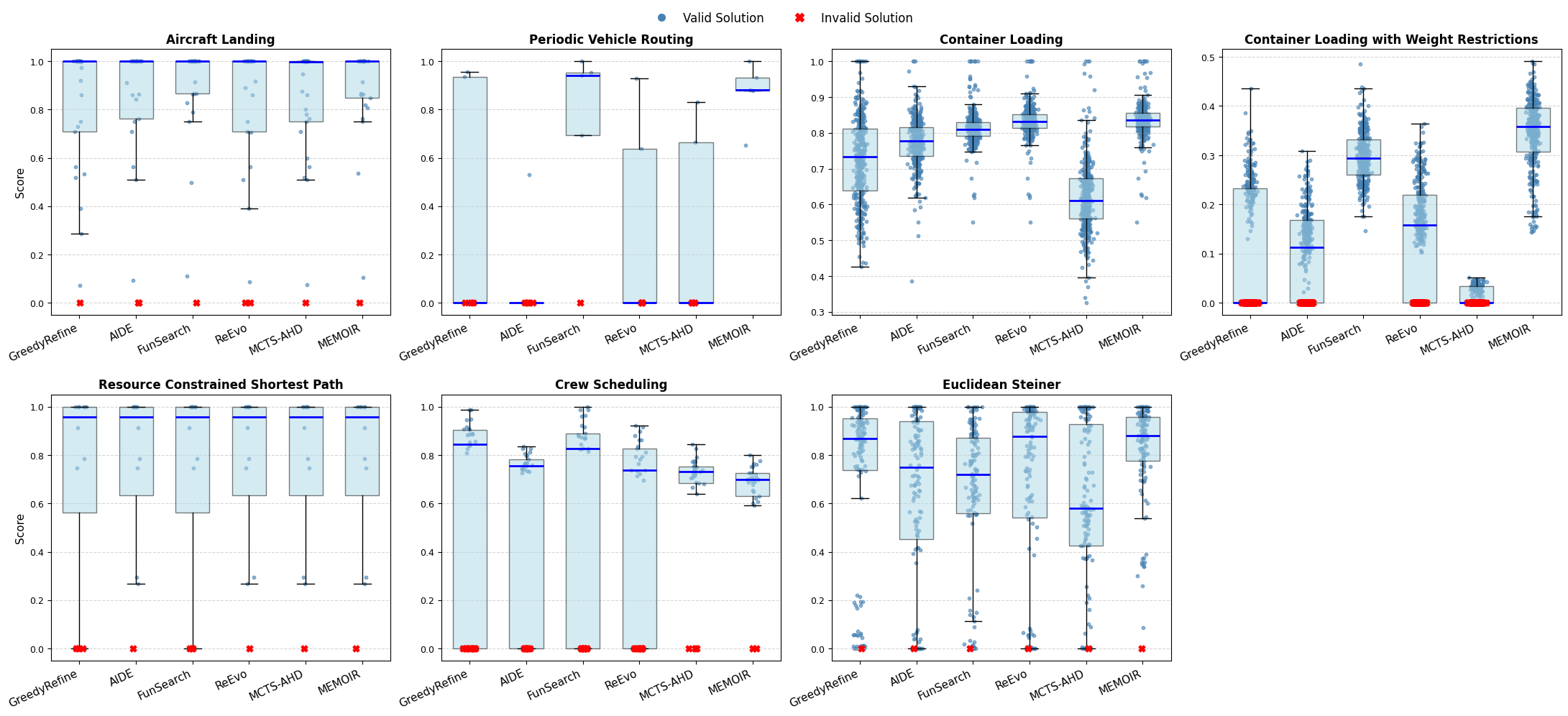}
  \vspace{-5mm}
  \caption{\textbf{Per-instance test-set score distributions across the seven CO problems.} Red crosses mark invalid solutions ($\text{score}{=}0$).}
  \label{fig:boxplots}
  \vspace{-7.5mm}
\end{figure*}

\begin{figure*}[b]
  \centering
  \vspace{-2em}
  \includegraphics[width=0.9\textwidth]{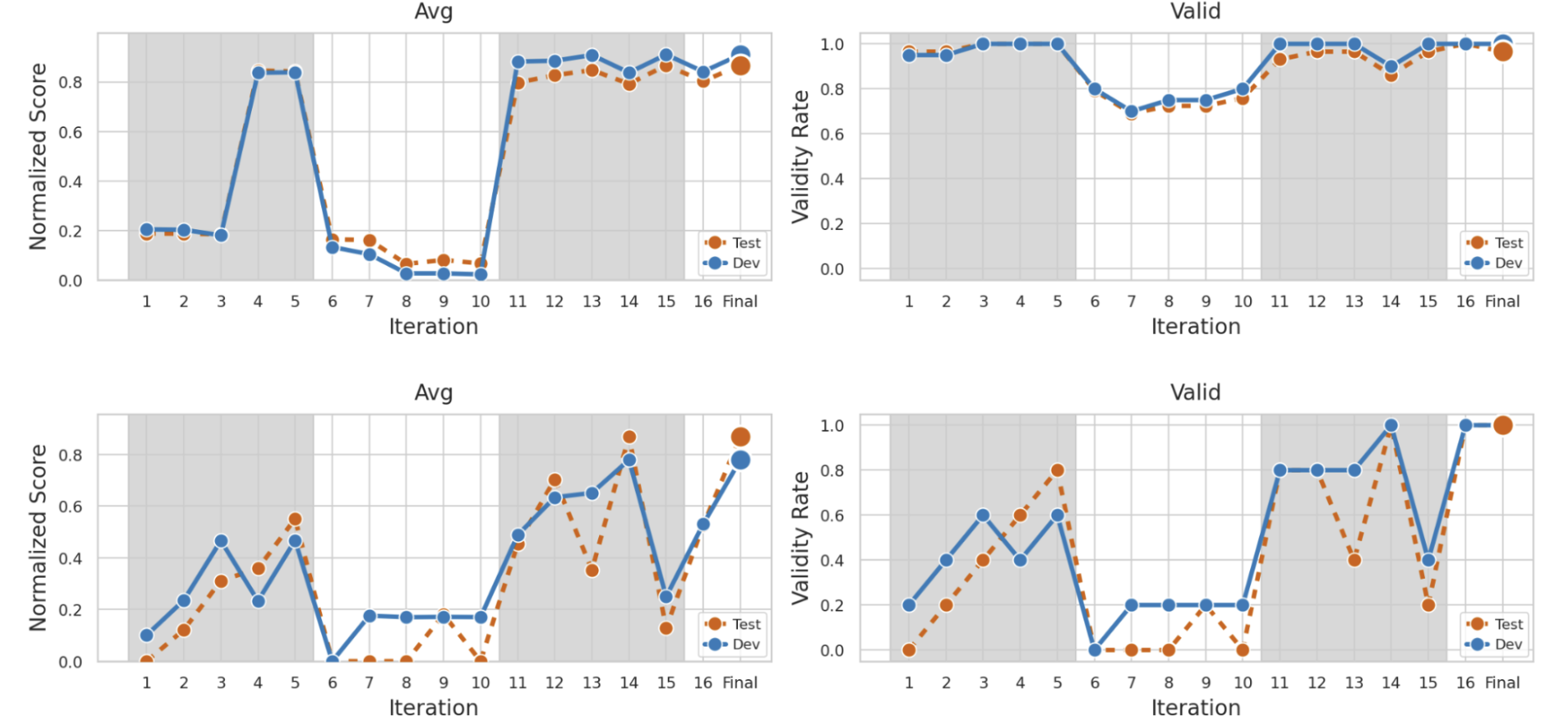}
  \vspace{-0.6em}
  \caption{\textbf{Convergence on the development set} for MEMOIR (\texttt{gpt-5-mini} w/ \texttt{gpt-5} critic). Avg (left) and Valid (right) over the $B{=}16$ executions on Aircraft Landing (top) and Periodic Vehicle Routing (bottom); alternating shaded regions mark branch boundaries. ``Final'' denotes the test-set score of the selected solver.}
  \label{fig:convergence}
\end{figure*}

\textbf{Overall performance.} Across seven domains (Table~\ref{tab:main_results}), MEMOIR (\texttt{gpt-5-mini} w/ \texttt{gpt-5} critic) attains the highest overall Avg ($.7301$) and Valid ($.9673$). The five LLM-agent baselines span $[.5332, .6570]$ Avg and $[.7343, .8755]$ Valid; the strongest, FunSearch, trails by $7.31 / 9.18$ points, and the non-LLM Classical Solver achieves $.4304$ Avg. The single-backbone variant \emph{MEMOIR (GPT-5-mini)} also surpasses every baseline on both overall metrics ($.6884 / .9201$), suggesting the gain is not solely attributable to a stronger critic. MEMOIR leads or matches the best LLM-agent on Avg on five of the seven domains; on the remaining two, FunSearch edges Aircraft Landing on Avg ($.8788$ vs.\ $.8668$) at matched Valid and MCTS-AHD edges Crew Scheduling on Avg ($.6353$ vs.\ $.6240$) but at lower Valid ($.8621$ vs.\ $.8966$), so no baseline dominates MEMOIR on both metrics in any single domain. Figure~\ref{fig:boxplots} corroborates this picture, with per-instance score distributions that are generally higher, and often tighter, than the baselines and with fewer infeasible outputs, and Figure~\ref{fig:convergence} shows within-branch improvement and later branches typically reaching higher peaks than earlier ones on Aircraft Landing and Periodic Vehicle Routing, with development and test curves tracking each other.

\textbf{MEMOIR delivers consistent gains across backbones.} Direct synthesis without an agent loop is unreliable on combinatorial structure: the best direct-synthesis backbone (o3-mini-high) reaches only $.4481$ overall Avg. Embedding any backbone in MEMOIR's full pipeline consistently lifts both metrics; with \texttt{gpt-5-mini}, this yields $+30.44$ Avg and $+24.17$ Valid over matched direct synthesis. The absolute level depends on the backbone's underlying coding capability, since solver synthesis for constraint-heavy CO problems is a demanding programming task, but the loop adds value at every capability tier we tested: notably, MEMOIR (Qwen) overtakes the open-source reasoning baselines DeepSeek-R1-Distill-Llama-70B and QwQ-32B on both overall metrics, showing that structured memory makes a mid-tier coding backbone competitive with stronger reasoning models used without an agent loop.

\textbf{Memory quality and hierarchy both matter.} Two design checks isolate the role of memory. \emph{(i)~Memory entry quality matters:} Critic produces the diagnostic field of each branch-local record and Reflect produces global entries, so upgrading only those two from \texttt{gpt-5-mini} to \texttt{gpt-5} (moving from MEMOIR (GPT-5-mini) to MEMOIR (\texttt{gpt-5-mini} w/ \texttt{gpt-5} critic), with solver generation fixed to \texttt{gpt-5-mini} in both) lifts overall Avg by $+4.17$ and Valid by $+4.72$ ($.6884 / .9201 \rightarrow .7301 / .9673$), so the gain comes from sharper diagnostics and reflections yielding better-grounded memory. \emph{(ii)~Hierarchy itself matters:} on MEMOIR (\texttt{gpt-5-mini} w/ \texttt{gpt-5} critic), a flat-memory variant that collapses branch-local and global memory into a single shared memory drops Avg by $8.5$ points and Valid by $4.4$ points relative to the full method, indicating that the branch-local/global separation, not merely the presence of memory, drives the gains.

\textbf{Gains are stable across runs and instance difficulty.}
\label{sec:complexity}
Two checks probe robustness. \emph{(i)~Run-to-run stability:} across three independent runs on four domains (Aircraft Landing, Periodic Vehicle Routing, Container Loading, Resource Constrained Shortest Path; one representative per problem family), MEMOIR has the lowest variability among LLM-agent methods (Avg stdev $0.0098$, Valid stdev $0.0005$, arithmetic means across the four domains; full per-method table in Appendix~\ref{app:stability}). \emph{(ii)~Difficulty stratification:} stratifying \emph{test} instances into ordinal difficulty bins for Aircraft Landing and Periodic Vehicle Routing using instance-derived proxies (Appendix~\ref{app:complexity_proxies}; per-bin distributions in Appendix~\ref{app:complexity_results}), MEMOIR remains competitive with FunSearch across all Aircraft Landing bins, including the hardest, and sustains perfect validity across every Periodic Vehicle Routing bin, indicating that the improvements persist as constraints tighten.


\subsection{Ablation Study}

\begin{table*}[t]
  \caption{\textbf{Ablation of MEMOIR's memory components.} Average normalized score (Avg) and validity rate (Valid) across the seven CO problems.}
  \label{tab:ablation_detailed}
  \centering
  \begin{adjustbox}{max width=\textwidth}
  \begin{small}
  \setlength{\tabcolsep}{3pt}
  \renewcommand{\arraystretch}{1.40}

  \begin{tabular}{l
    S[table-format=1.4,table-column-width=\numw] S[table-format=1.4,table-column-width=\numw]
    S[table-format=1.4,table-column-width=\numw] S[table-format=1.4,table-column-width=\numw]
    S[table-format=1.4,table-column-width=\numw] S[table-format=1.4,table-column-width=\numw]
    S[table-format=1.4,table-column-width=\numw] S[table-format=1.4,table-column-width=\numw]
    S[table-format=1.4,table-column-width=\numw] S[table-format=1.4,table-column-width=\numw]
    S[table-format=1.4,table-column-width=\numw] S[table-format=1.4,table-column-width=\numw]
    S[table-format=1.4,table-column-width=\numw] S[table-format=1.4,table-column-width=\numw]
    S[table-format=1.4,table-column-width=\numw] S[table-format=1.4,table-column-width=\numw]
  }
    \toprule
    \textbf{Dataset} &
    \multicolumn{2}{c}{\dshead{Aircraft\\Landing}} &
    \multicolumn{2}{c}{\dshead{Periodic Vehicle \\ Routing}} &
    \multicolumn{2}{c}{\dshead{Container\\Loading}} &
    \multicolumn{2}{c}{\dshead{Container Loading \\ with Weight Restrictions}} &
    \multicolumn{2}{c}{\dshead{Resource Constrained\\Shortest Path}} &
    \multicolumn{2}{c}{\dshead{Crew\\Scheduling}} &
    \multicolumn{2}{c}{\dshead{Euclidean\\Steiner}} &
    \multicolumn{2}{c}{\dshead{Overall}} \\
    \midrule
    \textbf{Method}
    & \textbf{Avg} & \textbf{Valid}
    & \textbf{Avg} & \textbf{Valid}
    & \textbf{Avg} & \textbf{Valid}
    & \textbf{Avg} & \textbf{Valid}
    & \textbf{Avg} & \textbf{Valid}
    & \textbf{Avg} & \textbf{Valid}
    & \textbf{Avg} & \textbf{Valid}
    & \textbf{Avg} & \textbf{Valid} \\

    \cmidrule(lr){2-3}\cmidrule(lr){4-5}\cmidrule(lr){6-7}\cmidrule(lr){8-9}%
    \cmidrule(lr){10-11}\cmidrule(lr){12-13}\cmidrule(lr){14-15}\cmidrule(lr){16-17}

    MEMOIR (GPT-5-mini w/ GPT-5 critic)
      & .8668 & .9655
      & .8690 & 1.0000
      & .8364 & 1.0000
      & .3453 & 1.0000
      & .7508 & .9167
      & .6240 & .8966
      & .8183 & .9922
      & .7301 & .9673 \\
    \addlinespace[4pt]

    MEMOIR w/o Global Memory
      & .8623 & .9655
      & .5196 & .8000
      & .7914 & 1.0000
      & .3578 & 1.0000
      & .7508 & .9167
      & .5754 & .7241
      & .7765 & .9922
      & .6620 & .9141 \\
    \addlinespace[4pt]

    MEMOIR w/o Branch-Local Memory
      & .8263 & .9655
      & .4802 & .6000
      & .8094 & 1.0000
      & .1448 & .4143
      & .7508 & .9167
      & .5983 & .7241
      & .6544 & .9922
      & .6092 & .8018 \\
    \addlinespace[4pt]

    MEMOIR w/o Failed Nodes
      & .8459 & .9655
      & .7328 & .8000
      & .7718 & 1.0000
      & .2935 & 1.0000
      & .7508 & .9167
      & .4552 & .6207
      & .6748 & .9922
      & .6464 & .8993 \\
      \addlinespace[4pt]


    \bottomrule
  \end{tabular}

  \end{small}
  \end{adjustbox}
  \vspace{-1em}
\end{table*}

We ablate MEMOIR's memory components while keeping search procedure, prompts, models, and execution budget fixed, evaluating three variants in Table~\ref{tab:ablation_detailed}: (i) \emph{w/o Global Memory}, which removes global memory and reflection so branches initialize independently; (ii) \emph{w/o Branch-Local Memory}, which disables branch-local memory so refinement has no access to prior modifications or execution feedback within a branch; and (iii) \emph{w/o Failed Nodes}, which retains branch-local memory but stores only valid records, excluding invalid solvers and their diagnostics so the loop cannot avoid repeating previously encountered failures.

Removing global memory (\emph{w/o Global Memory}) reduces overall Avg by $6.81$ points and Valid by $5.32$ points. Disabling branch-local memory (\emph{w/o Branch-Local Memory}) causes the largest drop (Avg: $-12.09$, Valid: $-16.55$), with validity collapsing on Container Loading with Weight Restrictions ($1.0000 \rightarrow 0.4143$): refinement without execution history cannot reliably recover from constraint violations. Restricting branch-local memory to valid records (\emph{w/o Failed Nodes}) reduces Avg by $8.37$ and Valid by $6.80$ points, confirming that failure information is essential for avoiding repeated infeasible designs.
\vspace{-0.8em}




\section{Conclusion}
\vspace{-0.5em}
We introduced MEMOIR, a memory-guided tree-search framework for LLM-based solver synthesis whose two-level memory hierarchy separates branch-local execution-grounded refinement from compressed cross-branch algorithmic summaries, with a reflection step that distills each branch into transferable lessons before they enter global memory. Across seven classical CO problems under a fixed 16-execution budget, MEMOIR reaches 96.7\% validity (a 9.2-point gap over the strongest LLM baseline) and improves average normalized score by 7.3 points, with run-to-run validity standard deviation over an order of magnitude below every baseline. Ablations attribute the gains to all three components: removing branch-local memory hurts most, while excluding failure records or removing global memory each causes meaningful drops. The broader takeaway is that separating transferable algorithmic insight from low-level execution detail is an effective design principle for LLM-based search agents, particularly where infeasibility is the primary bottleneck.

\vspace{-0.8em}

\paragraph{Limitations and Future Work.}
\label{sec:limitations}
LLM-agent methods are evaluated only at the matched low budget $B{=}16$, which isolates search efficiency but does not characterize behavior at evolutionary-scale budgets. The rule that splits the budget across branches is heuristic rather than learned. Beyond search efficiency, our seven problems span scheduling, routing, packing, and geometric design, and MEMOIR treats each problem independently without transferring knowledge across problem types. Reflection itself relies on LLM summarization, which we mitigate by grounding global entries in the branch's execution trace and critic diagnostics rather than a free-form read of the code; even so, a misrepresented failure mode could in principle rule out a useful heuristic family. Future work includes scaling to evolutionary-scale budgets, replacing the heuristic branch-budget rule with a learned policy, persisting $\mathcal{M}_{\text{global}}$ across structurally related problems, hierarchical memory at multiple abstraction levels, and automated validation of global memory entries to improve reflection robustness.
\vspace{-0.8em}
\section*{Impact Statement}
\vspace{-1em}
This work targets automated solver synthesis for combinatorial optimization, with potential to broaden access to high-quality solutions in domains such as logistics, healthcare scheduling, and resource allocation. As with any optimization tool, synthesized solvers can in principle be applied to objectives that conflict with societal values; MEMOIR is a tool for well-specified problems, not a decision-making system, and the choice of objectives and constraints rests with domain experts. MEMOIR also inherits the compute and energy costs of LLM inference; by avoiding repeated failures and redundant exploration of similar algorithmic families within a fixed budget, our memory-guided design uses each execution more productively, supporting stronger solver quality without scaling up compute.

\bibliographystyle{plainnat}
\bibliography{references}

\newpage
\appendix

\section{Problem Domains}
\label{app:datasets}

\subsection{Aircraft Landing}
\label{app:aircraft_landing}

\begin{problembox}
The problem is to schedule landing times for a set of planes across one or more runways such that each landing occurs within its prescribed time window and all pairwise separation requirements are satisfied; specifically, if plane i lands at or before plane j on the same runway, then the gap between their landing times must be at least the specified separation time provided in the input. In a multiple-runway setting, each plane must also be assigned to one runway, and if planes land on different runways, the separation requirement (which may differ) is applied accordingly. Each plane has an earliest, target, and latest landing time, with penalties incurred proportionally for landing before (earliness) or after (lateness) its target time. The objective is to minimize the total penalty cost while ensuring that no constraints are violated---if any constraint is breached, the solution receives no score.
\end{problembox}

\paragraph{Inputs}
An instance provides \texttt{num\_planes} and \texttt{num\_runways}. For each plane, \texttt{planes} contains its time window (\texttt{earliest}, \texttt{latest}), its preferred time (\texttt{target}), and its penalty rates (\texttt{penalty\_early}, \texttt{penalty\_late}). The instance also includes a matrix \texttt{separation} where \texttt{separation[i][j]} specifies the minimum required time gap if plane $i$ lands before plane $j$ on the same runway.

\paragraph{Outputs (solution format)}
The solver returns a dictionary \texttt{schedule} with one entry per plane. For each plane, the entry is a small dictionary containing its landing time and assigned runway.

\paragraph{Feasibility constraints}
A schedule is feasible only if every plane lands within its allowed window, and every runway assignment is an integer between $1$ and \texttt{num\_runways}. In addition, for any two different planes assigned to the same runway, if plane $i$ is scheduled no later than plane $j$, then their landing times must be separated by at least \texttt{separation[i][j]}. Any violation makes the solution infeasible and yields no score.

\paragraph{Objective value}
For each plane, landing earlier than \texttt{target} incurs a penalty proportional to how early it lands and \texttt{penalty\_early}; landing later incurs a penalty proportional to how late it lands and \texttt{penalty\_late}. The objective value returned by the evaluator is the total penalty summed over all planes, and the goal is to minimize this total.


\subsection{Periodic Vehicle Routing}
\label{app:period_vehicle_routing}

\begin{problembox}
The Periodic Vehicle Routing Problem requires planning delivery routes over a multi-day planning period.

Each customer (other than the depot, whose id is 0) is provided with a list of candidate service schedules. A schedule is represented by a binary vector of length equal to the period (e.g., [1, 0, 1] for a 3-day period), where a 1 in a given position indicates that the customer must be visited on that day. The decision maker must select exactly one candidate schedule for each customer.

For every day in the planning period, if a customer's chosen schedule indicates a delivery (i.e., a 1),
then exactly one vehicle must visit that customer on that day. Otherwise, the customer should not be visited. The decision maker must also design, for each day, the tours for the vehicles. Each tour is a continuous route that starts at the depot (id 0) and, after visiting a subset of customers, returns to the depot. Each vehicle is only allowed to visit the depot once per day---namely, as its starting and ending point---and it is not allowed to return to the depot in the middle of a tour.

Moreover, each vehicle route must obey a capacity constraint: the total demand of the customers visited on that tour must not exceed the vehicle capacity each day. Although multiple vehicles are available per day (as specified by the input), not all available vehicles have to be used, but the number of tours in a given day cannot exceed the provided number of vehicles. In addition, the tours on each day must cover exactly those customers who require service per the selected schedules, and no customer may be visited more than once in a given day.

The objective is to choose a schedule for every customer and plan the daily tours so as to minimize the overall distance traveled
by all vehicles during the entire planning period. Distances are measured using Euclidean distance.
\end{problembox}

\paragraph{Inputs}
An instance provides the depot location \texttt{depot} (id 0) and a list of customers \texttt{customers}, each with coordinates, a demand \texttt{demand}, and a list of candidate service schedules \texttt{schedules}. It also provides the planning horizon \texttt{period\_length}, the number of vehicles available each day \texttt{vehicles\_per\_day} (one value per day), and the per-vehicle capacity \texttt{vehicle\_capacity}.

\paragraph{Outputs (solution format)}
The solver returns two objects. First, \texttt{selected\_schedules} maps each customer id to exactly one chosen schedule from that customer's candidate list. Second, \texttt{tours} maps each day to a list of tours, where each tour is a list of vertex ids that starts at the depot (0), visits some customers, and ends at the depot (0).

\paragraph{Feasibility constraints}
The solution is feasible only if each customer is assigned a chosen schedule that is exactly one of its candidates and has length \texttt{period\_length}. For each day, the set of customers visited across all tours must match exactly the customers whose chosen schedule has a $1$ on that day: every required customer is visited exactly once, and no other customer is visited. Each tour must start and end at the depot and may not include the depot in the middle of the route, and it may not visit the same customer more than once. Finally, each tour must respect the capacity limit: the sum of demands of customers on that tour must not exceed \texttt{vehicle\_capacity}. If any constraint is violated, the solution is infeasible and yields no score.

\paragraph{Objective value}
The evaluator computes the total travel distance over all tours on all days using Euclidean distances between consecutive vertices in each tour. The objective is to minimize this total distance.


\subsection{Container Loading}
\label{app:container_loading}

\begin{problembox}
Solves a container loading problem: Given a 3D container of specified dimensions and multiple box types---each defined by dimensions, orientation constraints, and available quantity---the goal is to optimally place these boxes within the container to maximize the volume utilization ratio. Each box placement must respect orientation constraints (vertical alignment flags), fit entirely within container boundaries, and avoid overlaps. The solution returns precise coordinates and orientations for each box placement, quantified by a volume utilization score calculated as the total volume of placed boxes divided by the container volume. Invalid placements result in a score of 0.0.
\end{problembox}

\paragraph{Inputs}
An instance provides the container dimensions \texttt{container} as (\texttt{container\_length}, \texttt{container\_width}, \texttt{container\_height}). It also provides a set of box types \texttt{box\_types}. For each box type, the input specifies its three side lengths \texttt{dims} = [d1,d2,d3], a binary vector \texttt{flags} = [f1,f2,f3] indicating which side length is allowed to be the vertical dimension, and an available quantity \texttt{count}. 

\paragraph{Outputs (solution format)}
The solver returns a dictionary with key \texttt{placements}. Each placement is one packed box and is represented by seven integers:
\texttt{box\_type}, \texttt{container\_id}, \texttt{x}, \texttt{y}, \texttt{z}, \texttt{v}, \texttt{hswap}.
Here (\texttt{x},\texttt{y},\texttt{z}) is the lower-left-bottom corner of the box inside the container; \texttt{v} selects which dimension index (0,1,2) is used as the vertical side length; and \texttt{hswap} indicates whether the remaining two horizontal side lengths are swapped.

\paragraph{Feasibility constraints}
A placement is feasible only if its chosen vertical index \texttt{v} is allowed (\texttt{flags[v]} = 1). Given \texttt{v}, the vertical side length is \texttt{dims[v]}; the other two side lengths form the horizontal footprint, and \texttt{hswap} determines their order. Each placed box must lie entirely within the container bounds (nonnegative coordinates and not exceeding the container length, width, and height). Any two placed boxes in the same \texttt{container\_id} must be non-overlapping (touching faces/edges is allowed). Finally, the number of placements of each box type may not exceed its available \texttt{count}. If any constraint is violated, the solution is invalid and receives score $0.0$.

\paragraph{Objective value}
If the packing is valid, the evaluator returns the volume utilization ratio: the total volume of all placed boxes divided by the container volume. The goal is to maximize this ratio.


\subsection{Container Loading with Weight Restrictions}
\label{app:container_loading_weight}

\begin{problembox}
The Container Loading with Weight Restrictions problem aims to maximize the utilization of a container's volume by selecting and strategically placing boxes inside it. Given a container with specified dimensions (length, width, height) and multiple types of boxes, each characterized by their dimensions, quantities, weights, and load-bearing constraints, the optimization goal is to determine the placement and orientation of these boxes (with each box allowed three possible orientations) that maximizes the ratio of total occupied box volume to container volume. The solution must strictly adhere to spatial constraints (boxes must fit entirely within the container without overlapping), load-bearing constraints (boxes must support the weight of boxes stacked above them according to given limits), and orientation restrictions. The optimization quality is evaluated by the achieved utilization metric, defined as the total volume of successfully placed boxes divided by the container volume; if any constraint is violated, the utilization score is zero.
\end{problembox}

\paragraph{Inputs}
An instance provides the container dimensions \texttt{container} as (L, W, H) in cm. It also provides \texttt{box\_types}, a list of box-type specifications. For each box type, the input includes its dimensions (\texttt{length}, \texttt{width}, \texttt{height}), binary orientation flags (\texttt{length\_flag}, \texttt{width\_flag}, \texttt{height\_flag}) that indicate which side is allowed to be vertical, an available quantity \texttt{count}, a box weight \texttt{weight}, and three load-bearing parameters \texttt{lb1}, \texttt{lb2}, \texttt{lb3} (one for each orientation choice). The fields \texttt{n}, \texttt{instance}, and \texttt{cargo\_vol} appear in the input interface (and \texttt{cargo\_vol} is not used in evaluation).

\paragraph{Outputs (solution format)}
The solver returns a list \texttt{placements} of placed boxes. Each placement specifies a box type (\texttt{box\_type}, 1-indexed), an orientation (\texttt{orientation} in \{1,2,3\}), and the position (\texttt{x}, \texttt{y}, \texttt{z}) of the lower-left-front corner in cm. 

\paragraph{Feasibility constraints}
Each placement must use a valid box type index and an allowed orientation (an orientation is allowed only if the corresponding vertical-flag is 1). Given the chosen orientation, the box has fixed axis-aligned dimensions inside the container. Every placed box must lie entirely within the container bounds, and the number of used boxes of each type may not exceed its available \texttt{count}. Boxes must not overlap in space (touching is allowed). Any box that is not on the floor must be fully supported from below by exactly one other box: its bottom face must sit exactly on the top face of a single supporting box, and its horizontal footprint must be contained within the supporter's top face. Finally, load-bearing constraints must hold: each supporting box must be able to carry the total weight of all boxes stacked above it (including indirect stacks), subject to its orientation-specific load-bearing limit. If any constraint is violated, the solution is invalid and receives score $0.0$.

\paragraph{Objective value}
If the packing is valid, the evaluator returns the utilization ratio: the total volume of all placed boxes (using their original dimensions) divided by the container volume. The goal is to maximize this utilization.


\subsection{Resource Constrained Shortest Path}
\label{app:rcsp}

\begin{problembox}
This problem involves finding the shortest path from vertex 1 to vertex n in a directed graph while satisfying resource constraints. Specifically, each vertex and arc has associated resource consumptions, and the cumulative consumption for each resource must fall within the provided \texttt{lower\_bounds} and \texttt{upper\_bounds}. The input includes the number of vertices (\texttt{n}), arcs (\texttt{m}), resource types (\texttt{K}), resource consumption at each vertex, and a graph represented as a mapping from vertices to lists of arcs (each arc being a tuple of end vertex, cost, and arc resource consumptions). The optimization objective is to minimize the total arc cost of the path, with the condition that the path is valid---meaning it starts at vertex 1, ends at vertex n, follows defined transitions in the graph, and respects all resource bounds; if any of these constraints are not met, the solution receives no score.
\end{problembox}

\paragraph{Inputs}
An instance provides the number of vertices \texttt{n}, the number of directed arcs \texttt{m}, and the number of resource types \texttt{K}. For each resource, it provides a lower bound \texttt{lower\_bounds} and an upper bound \texttt{upper\_bounds}. It also provides \texttt{vertex\_resources}, which gives the resource consumption of each vertex, and a directed graph \texttt{graph} that lists outgoing arcs for each vertex. Each arc is represented as \texttt{(end\_vertex, cost, arc\_resources)}, where \texttt{cost} is the arc cost and \texttt{arc\_resources} is a length-\texttt{K} list of resource consumptions on that arc.

\paragraph{Outputs (solution format)}
The solver returns a dictionary with a \texttt{path}, which is a list of vertex indices describing the chosen route from vertex 1 to vertex \texttt{n}. The output also includes a \texttt{total\_cost} field, but the evaluator recomputes the cost from the arcs along \texttt{path}.

\paragraph{Feasibility constraints}
A solution is feasible only if \texttt{path} starts at vertex 1 and ends at vertex \texttt{n}. Every consecutive pair of vertices in \texttt{path} must correspond to a directed arc that exists in \texttt{graph}. Resource consumption is accumulated by adding the resource vector of every visited vertex and also adding the resource vector of every traversed arc. For each resource type $k \in \{1,\dots,\texttt{K}\}$, the final total consumption must lie within the interval $[\texttt{lower\_bounds}[k],\,\texttt{upper\_bounds}[k]]$. If any of these checks fail, the solution is invalid and receives no score.

\paragraph{Objective value}
If the path is valid, the evaluator returns the total arc cost along the path, computed by summing the \texttt{cost} values of the arcs used. The goal is to minimize this total cost.


\subsection{Crew Scheduling}
\label{app:crew_scheduling}

\begin{problembox}
The Crew Scheduling Problem involves assigning each task---with defined start and finish times---to exactly one crew, aiming to minimize the total transition costs between consecutive tasks. Each crew's schedule must satisfy three constraints: tasks within a crew must not overlap; valid transitions (with associated costs) must exist between every consecutive pair of tasks; and the crew's total duty time (from the start of the first task to the finish of the last) cannot exceed a specified time limit. Additionally, no more than K crews can be used to cover all tasks. Solutions violating any of these constraints are considered infeasible and receive no score. The optimization objective is therefore to determine assignments of tasks to no more than K crews that minimize the sum of transition costs while strictly adhering to all constraints, yielding a feasible and cost-effective scheduling solution.
\end{problembox}

\paragraph{Inputs}
An instance provides the number of tasks \texttt{N}, the crew limit \texttt{K}, and a duty-time limit \texttt{time\_limit}. Task times are given by \texttt{tasks}, which maps each task id to its (\texttt{start\_time}, \texttt{finish\_time}). Transition information is given by \texttt{arcs}, which maps a pair (\texttt{from\_task}, \texttt{to\_task}) to a transition cost.

\paragraph{Outputs (solution format)}
The solver returns a list \texttt{crews}. Each element is an ordered list of task ids, representing the sequence of tasks assigned to one crew.

\paragraph{Feasibility constraints}
Every task must appear exactly once across all crew lists. The solution may use at most \texttt{K} crews (i.e., \texttt{len(crews)} $\le$ \texttt{K}), and no crew list may be empty. Within each crew, tasks must be in non-overlapping time order: the finish time of a task may not exceed the start time of the next task in the same crew. For every consecutive pair of tasks in a crew, there must be a valid transition arc in \texttt{arcs}. The duty time of each crew, defined as the finish time of its last task minus the start time of its first task, must not exceed \texttt{time\_limit}. If any constraint is violated, the solution is infeasible and receives no score.

\paragraph{Objective value}
If the assignment is feasible, the evaluator returns the total transition cost, computed by summing the costs of all consecutive task transitions across all crews. The goal is to minimize this total cost.


\subsection{Euclidean Steiner Tree}
\label{app:euclidean_steiner}

\begin{problembox}
Given a set of 2D points (terminals), the goal of the Euclidean Steiner Problem is to compute a tree connecting all terminals with minimum total length. The total length is measured as the sum of Euclidean distances (where the Euclidean distance between two points (x1,y1) and (x2,y2) is sqrt((x1-x2)² + (y1-y2)²)). Unlike a Minimum Spanning Tree (MST) computed solely on the given terminals, a Steiner tree may introduce extra points, called Steiner points, to reduce the overall length. In this formulation, it is assumed that the final candidate tree's total length is given by the MST computed on the union of the original terminals and the reported Steiner points. A lower ratio (candidate\_tree\_length/MST\_original\_length) indicates a better solution.
\end{problembox}

\paragraph{Inputs}
An instance provides \texttt{points}, a list of terminal coordinates. Each terminal is a pair \texttt{(x, y)} of floating-point values.

\paragraph{Outputs (solution format)}
The solver returns \texttt{steiner\_points}, a list of additional 2D points. Each Steiner point is also a pair \texttt{(x, y)}.

\paragraph{Feasibility constraints}
A solution is valid if the minimum spanning tree computed on the union of the terminals and the returned Steiner points is not longer than the minimum spanning tree computed on the terminals alone (within a small tolerance). If this check fails, the solution is invalid and receives no score.

\paragraph{Objective value}
Let \texttt{mst\_original} be the total length of the minimum spanning tree on the terminals, and let \texttt{candidate\_value} be the total length of the minimum spanning tree on the terminals plus the returned Steiner points. The evaluator computes the ratio \texttt{candidate\_value}/\texttt{mst\_original} and converts it into a score as $1-\texttt{candidate\_value}/\texttt{mst\_original}$. Smaller candidate length (and therefore a smaller ratio) yields a higher score.
\clearpage
\section{MEMOIR Procedure}
\label{app:algorithm}
Algorithm~\ref{alg:memoir} gives the full procedure introduced in Section~\ref{sec:memoir}. The five operators (\textsc{Propose}, \textsc{Repair}, \textsc{Improve}, \textsc{Critic}, \textsc{Reflect}) are realized by the prompts in Appendix~\ref{app:prompts}.

\begin{algorithm}[!h]
\caption{\textbf{MEMOIR: Memory-Guided Tree Search for Solver Synthesis.} Each branch is seeded by a proposal conditioned on global memory, refined by Repair/Improve with branch-local memory, and summarized by Reflect into a single global-memory entry; the highest-scoring valid solver on $\mathcal{D}_{\mathrm{dev}}$ is returned.}
\label{alg:memoir}
\small
\begin{algorithmic}[1]
\STATE \textbf{Input:} Problem description $P$; development set $\mathcal{D}_{\mathrm{dev}}$; test set $\mathcal{D}_{\mathrm{test}}$; execution budget $B$; per-branch depth cap $n$; per-instance timeout $T$.
\STATE \textbf{Initialize:} $\mathcal{M}_{\mathrm{global}} \leftarrow \emptyset$;\; $\mathcal{U} \leftarrow \emptyset$;\; $i \leftarrow 0$;\; remaining budget $b \leftarrow B$. \hfill\COMMENT{open new branches while $\geq 2$ executions remain}

\WHILE{$b \geq 2$}
    \STATE $i \leftarrow i + 1$
    \STATE \COMMENT{\textit{Exploration: initialize a new branch with a distinct algorithmic design}}
    \STATE $\mathcal{M}^{(i)}_{\mathrm{local}} \leftarrow \emptyset$
    \STATE $(s_0, \pi_0) \leftarrow \textsc{Propose}(P, \mathcal{M}_{\mathrm{global}})$
    \STATE $(v_0, f_0, E_0) \leftarrow \textsc{Execute}(s_0;\, \mathcal{D}_{\mathrm{dev}})$;\quad $\rho_0 \leftarrow \textsc{Critic}(s_0, \emptyset, E_0, \emptyset)$;\quad $b \leftarrow b - 1$
    \STATE $\mathcal{M}^{(i)}_{\mathrm{local}} \leftarrow \{(\pi_0, \rho_0, v_0, f_0, E_0)\}$

    \FOR{$t = 2$ to $n$}
        \STATE \COMMENT{\textit{Exploitation: extend the branch to depth $t$ via one repair or improve step; stop early when no further gain is expected or budget is spent}}
        \IF{$b = 0$ \textbf{or} $\neg\,\textsc{ImprovementExpected}(\mathcal{M}^{(i)}_{\mathrm{local}})$}
            \STATE \textbf{break}
        \ENDIF
        \IF{$v_u = 0$ for all $u \in \mathcal{M}^{(i)}_{\mathrm{local}}$}
            \STATE Sample invalid parent $u_{\mathrm{par}} \in \mathcal{M}^{(i)}_{\mathrm{local}}$ in proportion to $f_u$ (uniform if all $f_u = 0$)
            \STATE $(s', \pi') \leftarrow \textsc{Repair}(P, u_{\mathrm{par}}, \mathcal{M}^{(i)}_{\mathrm{local}})$
        \ELSE
            \STATE $u_{\mathrm{par}} \leftarrow \argmax_{u \in \mathcal{M}^{(i)}_{\mathrm{local}},\; v_u = 1} f_u$
            \STATE $(s', \pi') \leftarrow \textsc{Improve}(P, u_{\mathrm{par}}, \mathcal{M}^{(i)}_{\mathrm{local}})$
        \ENDIF
        \STATE $(v', f', E') \leftarrow \textsc{Execute}(s';\, \mathcal{D}_{\mathrm{dev}})$;\quad $\rho' \leftarrow \textsc{Critic}(s', s_{u_{\mathrm{par}}}, E', E_{u_{\mathrm{par}}})$;\quad $b \leftarrow b - 1$
        \STATE $\mathcal{M}^{(i)}_{\mathrm{local}} \leftarrow \mathcal{M}^{(i)}_{\mathrm{local}} \cup \{(\pi', \rho', v', f', E')\}$
    \ENDFOR

    \STATE \COMMENT{\textit{Cross-branch transfer: compress the branch into a single global-memory entry}}
    \STATE $g_i \leftarrow \textsc{Reflect}(\mathcal{M}^{(i)}_{\mathrm{local}})$
    \STATE $\mathcal{M}_{\mathrm{global}} \leftarrow \mathcal{M}_{\mathrm{global}} \cup \{g_i\}$;\quad $\mathcal{U} \leftarrow \mathcal{U} \cup \mathcal{M}^{(i)}_{\mathrm{local}}$
\ENDWHILE

\STATE \COMMENT{\textit{Final selection on dev (fall back to highest-scoring partial record if no fully valid solver was found); one-shot test evaluation}}
\IF{$\{u \in \mathcal{U} : v_u = 1\} \neq \emptyset$}
    \STATE $u^\star \leftarrow \argmax_{u \in \mathcal{U},\; v_u = 1} f_u$
\ELSE
    \STATE $u^\star \leftarrow \argmax_{u \in \mathcal{U}} f_u$
\ENDIF
\STATE $\hat s \leftarrow s_{u^\star}$
\STATE \textbf{Return} $\hat s$ and report $\bar f(\hat s; \mathcal{D}_{\mathrm{test}})$, $\textsc{Valid}(\hat s; \mathcal{D}_{\mathrm{test}})$
\end{algorithmic}
\end{algorithm}

\section{Prompt Templates}
\label{app:prompts}

This section lists the prompt templates used by MEMOIR's core components. 

\begin{promptbox}{Algorithm Proposer Prompt}
\small\ttfamily
You are an expert in Operation Research problem and combinatorial optimization solving a combinatorial optimization problem. In order to solve this problem more effectively, you need to come up with an excellent and creative algorithm for a solution and then implement this solution in Python. We will now provide a description of the problem. \\

\textbf{Task description} \\
\textit{[CO-Bench problem description goes here]} \\

\textbf{Past Failures \& Constraints} \\
\textit{[Global memory summary goes here]} \\

\textbf{Instructions}

\textbf{Response format} \\
Your response should be a brief outline/sketch of your proposed solution in natural language (3--5 sentences), followed by a single markdown code block (wrapped in \texttt{```}) which implements this solve function and yields the required formatted output. There should be no additional headings or text in your response. Just natural language text followed by a newline and then the markdown code block. \\

\textbf{Solution sketch guideline} \\
This is a NEW exploration phase. Review the Past Failures \& Constraints section carefully. You MUST propose an algorithm design that DIFFERS from failed attempts and is NOT simply a repetition of previous successful ones. This first solution design should be relatively simple. The solution sketch should be 3--5 sentences. Enclose all your code within a code block: \texttt{``` ... ```} and name the main function \texttt{def solve(**kwargs)}. Your function has a 10-second timeout; aim to yield the best possible results within this limit. \\

\textbf{Implementation guideline} \\
The code should implement the proposed solution and yield solutions over time as in the template solve function. Use Python's \texttt{yield} keyword repeatedly to produce a stream of solutions. Each yielded solution should be better than the previous one. You can end the generator early by using \texttt{return} if you've found an optimal solution. The evaluation will be performed on the last solution you yielded before timeout. Note that you don't need to handle timeout --- the system will automatically use the last solution you yielded before timeout. If your function doesn't yield any solution before timeout, it will be considered a timeout failure. The code should be a single-file python program that is self-contained and can be executed as-is. No parts of the code should be skipped, don't terminate before finishing the script. Your response should only contain a single code block. Be aware of the running time of the code, it should complete within 10 seconds. You only need to implement the \texttt{def solve(**kwargs)} function. The data loading and evaluation are all done in the grading system. \\

\textbf{Installed Packages} \\
You may use any standard Python libraries and general-purpose packages (e.g., numpy, pandas); they are already installed. Do NOT use any optimization or solver libraries, including ortools, gurobi, pulp, pyomo, cvxpy, mip, z3, scip, cplex, or similar.
\end{promptbox}


\begin{promptbox}{Improve Prompt}
\small\ttfamily
You are an expert in Operation Research problem and combinatorial optimization, solving a combinatorial optimization problem. You are provided with a previously developed solution below and should improve it in order to further increase the performance of algorithm. For this you should first outline a brief plan in natural language for how the algorithm can be improved and then implement this improvement in Python based on the provided previous solution. \\

\textbf{Task description} \\
\textit{[CO-Bench problem description goes here]} \\

\textbf{Memory} \\
\textit{[Branch-local history for the current branch goes here]} \\ 

\textbf{Previous solution} \\
\textit{[Previous solver code goes here]} \\

\textbf{Instructions}

\textbf{Response format} \\
Your response should be a brief outline/sketch of your proposed solution in natural language (3--5 sentences), followed by a single markdown code block (wrapped in \texttt{```}) which implements this solve function and yields the required formatted output. There should be no additional headings or text in your response. Just natural language text followed by a newline and then the markdown code block. \\

\textbf{Solution improvement sketch guideline} \\
The solution sketch should be a brief natural language description of how the previous solution can be improved. Review the Memory section. Identify patterns in \texttt{No bugs, worsened score} or \texttt{No change}. DO NOT repeat those strategies. If the Memory shows a previously successful approach, you must SIGNIFICANTLY vary it; do not just re-apply the exact same logic. You should be very specific and should only propose a single actionable improvement. This improvement should be atomic. The solution sketch should be 3--5 sentences. \\

\textbf{Implementation guideline} \\
The code should implement the proposed solution and yield solutions over time as in the template solve function. Use Python's \texttt{yield} keyword repeatedly to produce a stream of solutions. Each yielded solution should be better than the previous one. You can end the generator early by using \texttt{return} if you've found an optimal solution. The evaluation will be performed on the last solution you yielded before timeout. Note that you don't need to handle timeout --- the system will automatically use the last solution you yielded before timeout. If your function doesn't yield any solution before timeout, it will be considered a timeout failure. The code should be a single-file python program that is self-contained and can be executed as-is. No parts of the code should be skipped, don't terminate before finishing the script. Your response should only contain a single code block. Be aware of the running time of the code, it should complete within 10 seconds. You only need to implement the \texttt{def solve(**kwargs)} function. The data loading and evaluation are all done in the grading system. \\

\textbf{Installed Packages} \\
You may use any standard Python libraries and general-purpose packages (e.g., numpy, pandas); they are already installed. Do NOT use any optimization or solver libraries, including ortools, gurobi, pulp, pyomo, cvxpy, mip, z3, scip, cplex, or similar.

\end{promptbox}


\begin{promptbox}{Debug Prompt}
\small\ttfamily
You are an expert in Operation Research problem and combinatorial optimization, solving a combinatorial optimization problem. Your previous solution had a bug, so based on the information below, you should revise it in order to fix this bug. Your response should be an implementation outline in natural language, followed by a single markdown code block which implements the bugfix/solution. \\

\textbf{Task description} \\
\textit{[CO-Bench problem description goes here]} \\

\textbf{Memory} \\
\textit{[Branch-local history for the current branch goes here]} \\

\textbf{Previous (buggy) implementation} \\
\textit{[Buggy solver code goes here]} \\

\textbf{Execution output} \\
\textit{[Runtime logs / constraint violations go here]} \\

\textbf{Instructions}

\textbf{Response format} \\
Your response should be a brief outline/sketch of your proposed solution in natural language (3--5 sentences), followed by a single markdown code block (wrapped in \texttt{```}) which implements this solve function and yields the required formatted output. There should be no additional headings or text in your response. Just natural language text followed by a newline and then the markdown code block. \\

\textbf{Bugfix improvement sketch guideline} \\
Review the Execution output and identify the first concrete failure (e.g., missing key, wrong return format, constraint violation, crash). If the Memory shows a similar bug or a previously attempted fix, do not repeat the same fix if it did not work; try a different approach. Focus on restoring correctness first: fix the bug with the smallest possible change, without redesigning the whole algorithm. The sketch should be 3--5 sentences. \\

\textbf{Implementation guideline} \\
The code should implement the proposed solution and yield solutions over time as in the template solve function. Use Python's \texttt{yield} keyword repeatedly to produce a stream of solutions. Each yielded solution should be better than the previous one. You can end the generator early by using \texttt{return} if you've found an optimal solution. The evaluation will be performed on the last solution you yielded before timeout. Note that you don't need to handle timeout --- the system will automatically use the last solution you yielded before timeout. If your function doesn't yield any solution before timeout, it will be considered a timeout failure. The code should be a single-file python program that is self-contained and can be executed as-is. No parts of the code should be skipped, don't terminate before finishing the script. Your response should only contain a single code block. Be aware of the running time of the code, it should complete within 10 seconds. You only need to implement the \texttt{def solve(**kwargs)} function. The data loading and evaluation are all done in the grading system. \\

\textbf{Installed Packages} \\
You may use any standard Python libraries and general-purpose packages (e.g., numpy, pandas); they are already installed. Do NOT use any optimization or solver libraries, including ortools, gurobi, pulp, pyomo, cvxpy, mip, z3, scip, cplex, or similar.
\end{promptbox}


\begin{promptbox}{Trajectory Critic Prompt}
\small\ttfamily
You are an expert in combinatorial optimization. You have written code to solve this task and now need to evaluate the output. You must compare the Previous Version Execution Output with the Current Execution Output. Determine if the changes fixed previous bugs, introduced new ones, or improved performance. \\

\textbf{Task description} \\
\textit{[CO-Bench problem description goes here]} \\

\textbf{Previous Version Implementation} \\
\textit{[Parent solver code goes here, or write: No previous implementation (Initial Draft)]} \\

\textbf{Current Implementation} \\
\textit{[Current solver code goes here]} \\

\textbf{Previous Version Execution Output} \\
\textit{[Previous logs go here, or write: No previous logs (Initial Draft)]} \\

\textbf{Current Version Execution Output} \\
\textit{[Current logs go here]} \\

\textbf{Instructions} \\
Return a JSON object with the following fields. \\

\texttt{is\_bug}: true if the output log shows that execution failed or has some bug, otherwise false. Note that timeout is not a bug; in that case set \texttt{is\_bug} to false. \\

\texttt{summary}: compare previous and current logs. If there is a bug, propose a fix. If fixed, explain what improved. If the score changed, mention the delta. Write 2--3 sentences. \\


Your response must contain only the JSON object.
\end{promptbox}


\begin{promptbox}{Reflection Prompt}
\small\ttfamily
Analyze the provided Trajectory History (a sequence of coding attempts) which failed to yield a satisfactory solution. Your goal is to extract a reusable lesson that prevents repeating the same suboptimal approach in the next fresh attempt.\\

\textbf{Trajectory History} \\
\textit{[Full branch-local memory for the finished branch goes here]} \\

\textbf{Instructions} \\
Return a JSON object with exactly the following keys. \\

\texttt{algorithmic design}: identify the core algorithmic approach that was attempted and refined throughout this branch. \\

\texttt{failure and stagnation reason}: explain the fundamental reason this approach failed or plateaued. \\

\texttt{constraint}: write a negative constraint (what NOT to do) that forces the next draft to take a meaningfully different approach.

Your response must contain only the JSON object.
\end{promptbox}

\section{Performance by Instance Complexity}
\label{app:complexity_proxies}

This appendix specifies the instance-derived difficulty proxies used to stratify test instances in Section~\ref{sec:main_results}. For each domain, we define a scalar hardness score $h(x) \geq 0$ computed from the instance data alone (no solver execution involved) and partition the test set into $K{=}3$ ordinal bins by the empirical terciles of $\{h(x) : x \in \mathcal{D}_{\text{test}}\}$. Because $h$ depends only on the instance, the binning is independent of any solver, the synthesis budget, or the method under evaluation.

\subsection{Aircraft Landing: Separation Pressure}
Aircraft Landing becomes harder when more aircraft share each runway (congestion) and required separations are large relative to available time-window slack (tightness). For an instance $x$, let $P$ be the number of planes and $R$ the number of runways. Each plane $i$ has an allowed window $[e_i,\ell_i]$; define the mean window width
\[
\overline{w} \;=\; \frac{1}{P}\sum_{i=1}^{P}(\ell_i-e_i).
\]
Let $\mathrm{sep}_{90}$ denote the 90th percentile of the off-diagonal entries of the pairwise separation matrix. We define the \emph{separation pressure} hardness score as
\[
h_{\textsc{AL}}(x)
\;=\;
\underbrace{\frac{P}{R}}_{\text{congestion}}
\cdot
\underbrace{\frac{\mathrm{sep}_{90}}{\overline{w}}}_{\text{relative tightness}}.
\]
Intuitively, $h_{\textsc{AL}}(x)$ increases when each runway must accommodate more planes and when typical required separations consume a larger fraction of the feasible time windows.

\subsection{Periodic Vehicle Routing: Balanced Peak Load Ratio}
Periodic Vehicle Routing becomes harder as total demand approaches daily capacity, especially because feasibility depends on selecting one service schedule per customer. For an instance $x$ with $n$ customers over $D$ days, each customer $i$ has demand $q_i$ and a set of candidate binary schedules $\mathcal{S}_i \subseteq \{0,1\}^{D}$ indicating the service days. Let $V_d$ be the number of vehicles available on day $d$ and $\mathrm{cap}$ be the per-vehicle capacity, so daily capacity is $C_d = V_d \cdot \mathrm{cap}$.
For any schedule selection $(s_i)_{i=1}^n$ with $s_i \in \mathcal{S}_i$, the induced demand on day $d$ is
\[
L_d \;=\; \sum_{i=1}^{n} q_i\, s_{i,d}.
\]
We define the \emph{balanced peak load ratio} hardness score as the best-achievable peak utilization over days:
\[
h_{\textsc{PRVP}}(x)
\;=\;
\min_{s_i \in \mathcal{S}_i}\;
\max_{d \in \{1,\dots,D\}}
\frac{L_d}{C_d}.
\]
Larger $h_{\textsc{PRVP}}(x)$ indicates tighter capacity.

\subsection{Per-Bin Score Distributions}
\label{app:complexity_results}

\begin{figure*}[t]
  \centering
  \begin{subfigure}[t]{0.49\linewidth}
    \centering
    \includegraphics[width=\linewidth]{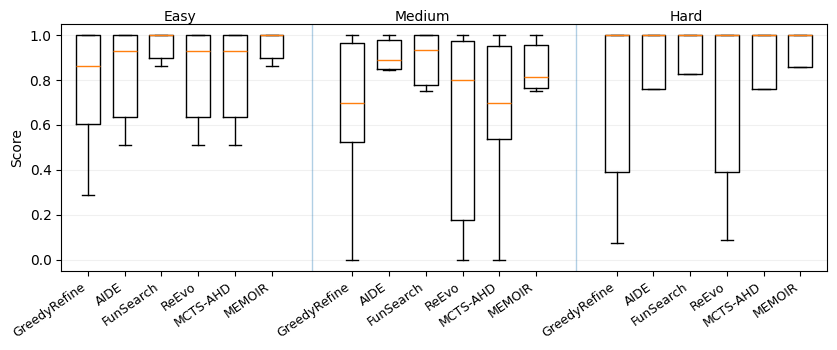}
    \caption{\textbf{Aircraft Landing}, stratified by separation pressure (runway congestion and separation tightness relative to time windows).}
    \label{fig:complexity_aircraft}
  \end{subfigure}
  \hfill
  \begin{subfigure}[t]{0.49\linewidth}
    \centering
    \includegraphics[width=\linewidth]{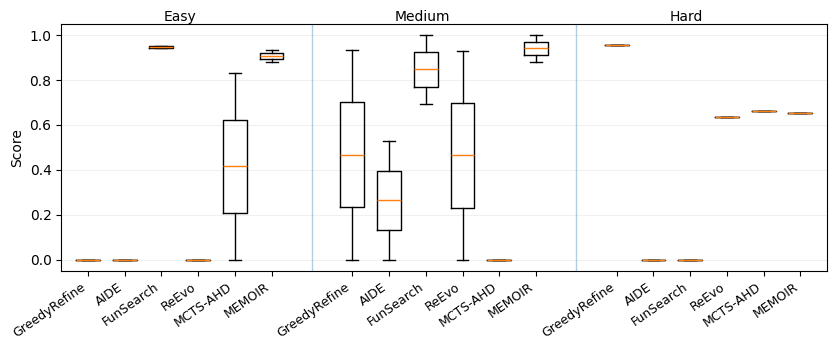}
    \caption{\textbf{Periodic Vehicle Routing}, stratified by balanced peak load ratio (demand relative to daily vehicle capacity).}
    \label{fig:complexity_prvp}
  \end{subfigure}
  \caption{\textbf{Performance by instance difficulty.} Test-set score distributions on each domain stratified into $K{=}3$ ordinal bins by an instance-only hardness proxy (\Cref{app:complexity_proxies}).}
  \label{fig:complexity}
  \vskip -0.10in
\end{figure*}

Using the proxies above, we partition each domain's test set into $K = 3$ ordinal bins and report per-bin score distributions in \Cref{fig:complexity_aircraft,fig:complexity_prvp}. For Aircraft Landing, MEMOIR matches or surpasses FunSearch on the hardest bin, so the small aggregate-Avg gap reported in Table~\ref{tab:main_results} reflects easy-instance variation rather than degradation under tight separations. For Periodic Vehicle Routing, MEMOIR sustains perfect validity across all bins while remaining competitive in score; GreedyRefine occasionally achieves higher Avg on hard instances but collapses to near-zero validity on easy and medium bins, and FunSearch produces infeasible solutions on the hardest bin.

\section{Token Consumption and Synthesis Cost Analysis}
\label{app:cost}

This appendix accounts for LLM token consumption and dollar cost of a synthesis run. All numbers reflect the \emph{one-time} cost of synthesizing a solver on a CO problem under the shared execution budget $B=16$; once synthesized, MEMOIR and every baseline emit a self-contained Python solver that is evaluated on every test instance with zero further LLM calls.

For each (method, problem) pair we report input and output token counts and the corresponding dollar cost. All methods use \texttt{gpt-5-mini} for solver code generation (Propose, Repair, Improve). \emph{MEMOIR (GPT-5-mini w/ GPT-5 critic)} and AIDE use \texttt{gpt-5} for their diagnostic operators (Critic and Reflect for MEMOIR, execution analysis for AIDE), while the single-backbone variant \emph{MEMOIR (GPT-5-mini)} runs all five operators on \texttt{gpt-5-mini}.

Across methods, synthesis cost stays under \$1 per run on every problem with wall-clock times of $\sim20$--$40$ minutes (Table~\ref{tab:cost_summary}), so differences are measured in cents. MEMOIR (\texttt{gpt-5-mini} w/ \texttt{gpt-5} critic) and AIDE share a similar cost envelope ($\sim\$0.52$ and $\sim\$0.53$ per run), reflecting their shared use of a \texttt{gpt-5} diagnostic operator. The single-backbone MEMOIR (\texttt{gpt-5-mini}) variant costs \$0.06 per run on average, placing it among the cheapest methods while running the full memory-guided pipeline.

\begin{table*}[h]
\centering
\caption{\textbf{Method-level token and time summary, averaged across the seven CO problems.} Input and output tokens are reported per synthesis run (in thousands); time is wall-clock minutes per run.}
\label{tab:cost_summary}
\begin{adjustbox}{max width=\textwidth}
\begin{small}
\setlength{\tabcolsep}{6pt}
\renewcommand{\arraystretch}{1.25}
\begin{tabular}{lccccccc}
\toprule
& \textbf{GreedyRefine} & \textbf{AIDE} & \textbf{FunSearch} & \shortstack{\textbf{MEMOIR}\\\textbf{(w/ GPT-5 critic)}} & \textbf{ReEvo} & \textbf{MCTS-AHD} & \shortstack{\textbf{MEMOIR}\\\textbf{(GPT-5-mini)}} \\
\midrule
Avg.\ Input Tokens (K)  & 110.6 & 159.5 & 139.3 & 168.6 & 467.7 & 164.5 & 178.6 \\
Avg.\ Output Tokens (K) &  94.2 &  57.5 &  74.1 &  62.6 &  76.4 &  87.1 &  60.6 \\
Avg.\ Cost (\$)         & 0.07  & 0.53  & 0.07  & 0.52  & 0.12  & 0.08  & 0.06  \\
Avg.\ Synthesis Time (min) & 29 & 33 & 27 & 36 & 21 & 24 & 27 \\
\bottomrule
\end{tabular}
\end{small}
\end{adjustbox}
\end{table*}

\section{Stability Across Runs}
\label{app:stability}

Table~\ref{tab:stability} reports the per-method run-to-run stability summarized in Section~\ref{sec:main_results}: arithmetic means of per-domain Avg and Valid standard deviation across three independent runs on the four representative problems listed in the caption. MEMOIR is markedly more consistent than every baseline.

\begin{table}[h]
  \caption{Standard deviation across 3 independent runs on four CO problems.}
  \label{tab:stability}
  \centering
  \begin{tabular}{lcc}
    \toprule
    \textbf{Method} & \textbf{Avg Stdev} & \textbf{Valid Stdev} \\
    \midrule
    GreedyRefine & 0.1043 & 0.1054 \\
    AIDE & 0.0295 & 0.0338  \\
    FunSearch & 0.0456 & 0.0579 \\
    ReEvo & 0.0973 & 0.0916 \\
    MCTS-AHD & 0.1349 & 0.1002 \\
    MEMOIR & 0.0098 & 0.0005 \\
    \bottomrule
  \end{tabular}
\end{table}

\section{Additional Search Dynamics Plots}
\begin{figure*}[!h]
  \centering
  \includegraphics[width=0.98\textwidth]{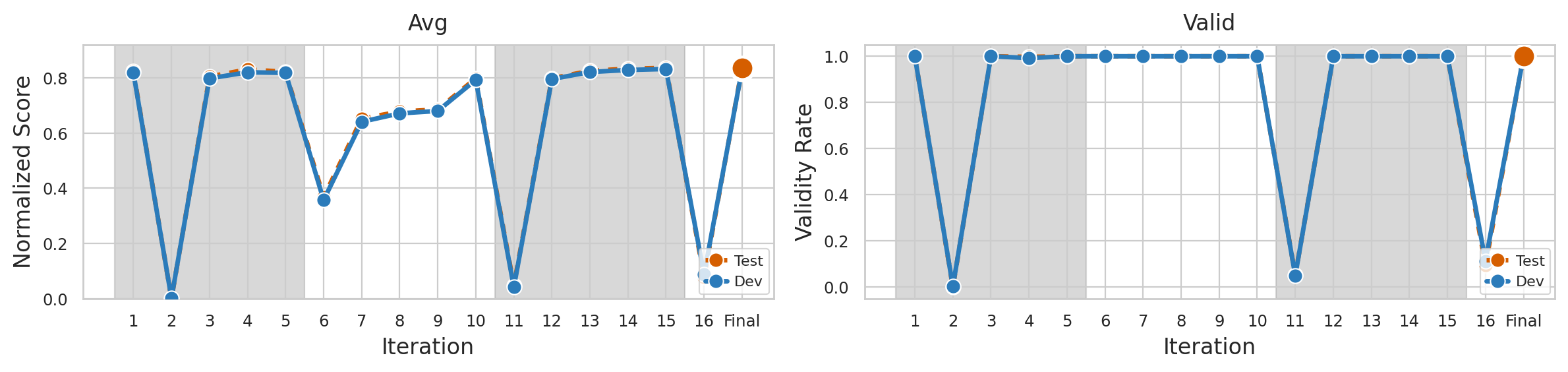}\\[-0.25em]
  \includegraphics[width=0.98\textwidth]{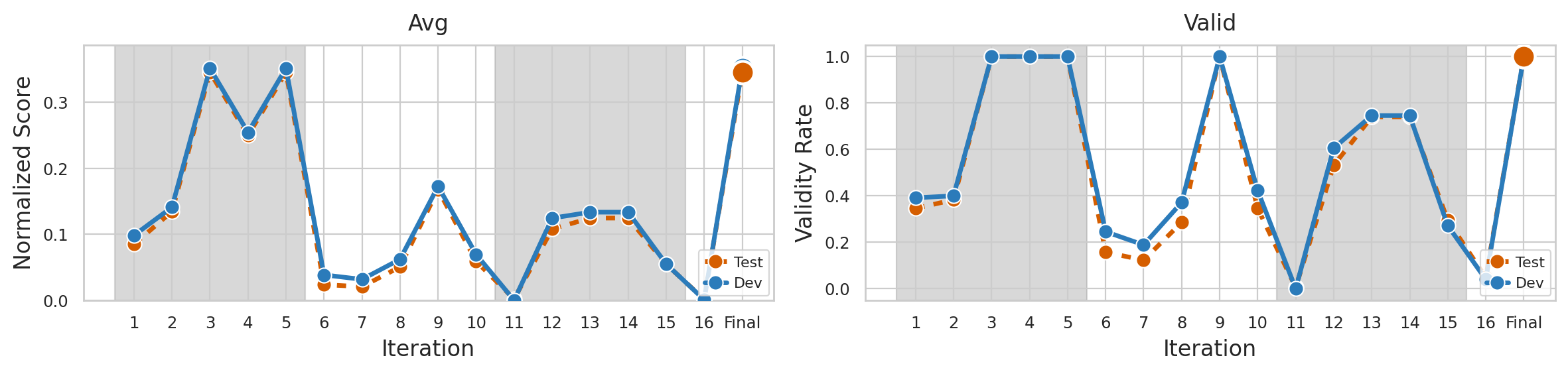}\\[-0.25em]
  \includegraphics[width=0.98\textwidth]{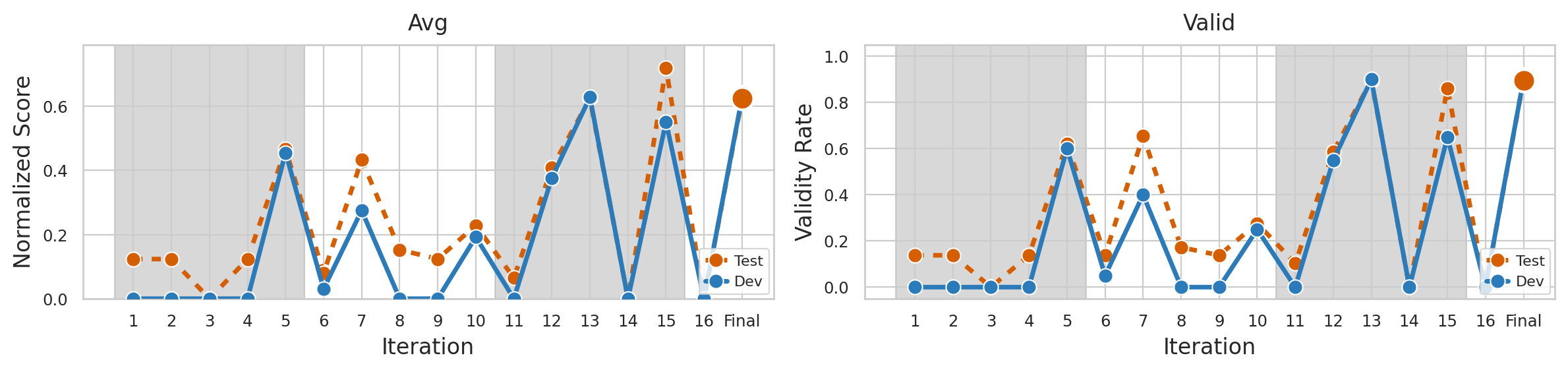}\\[-0.25em]
  \includegraphics[width=0.98\textwidth]{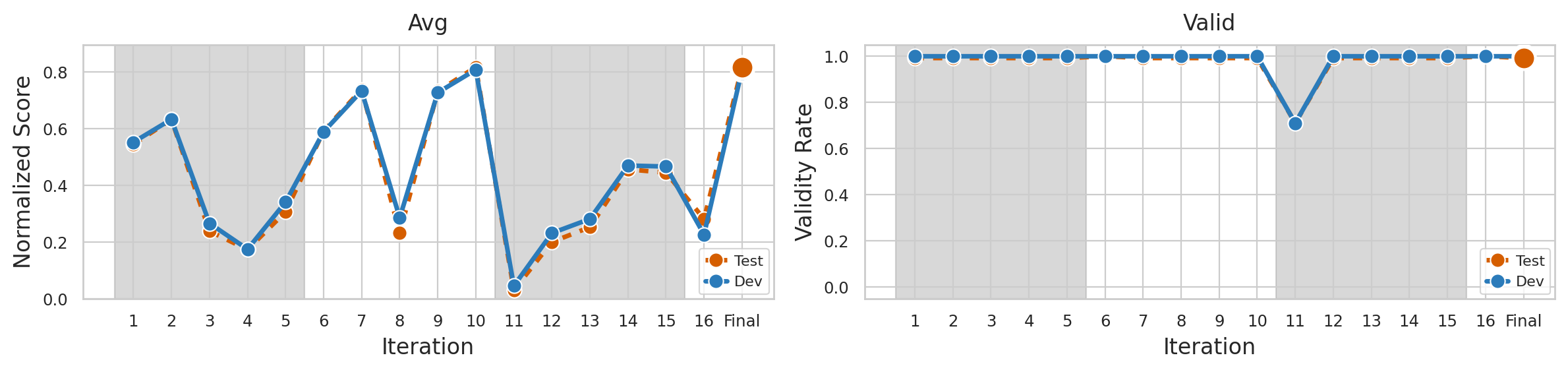}\\[-0.25em]
  \includegraphics[width=0.98\textwidth]{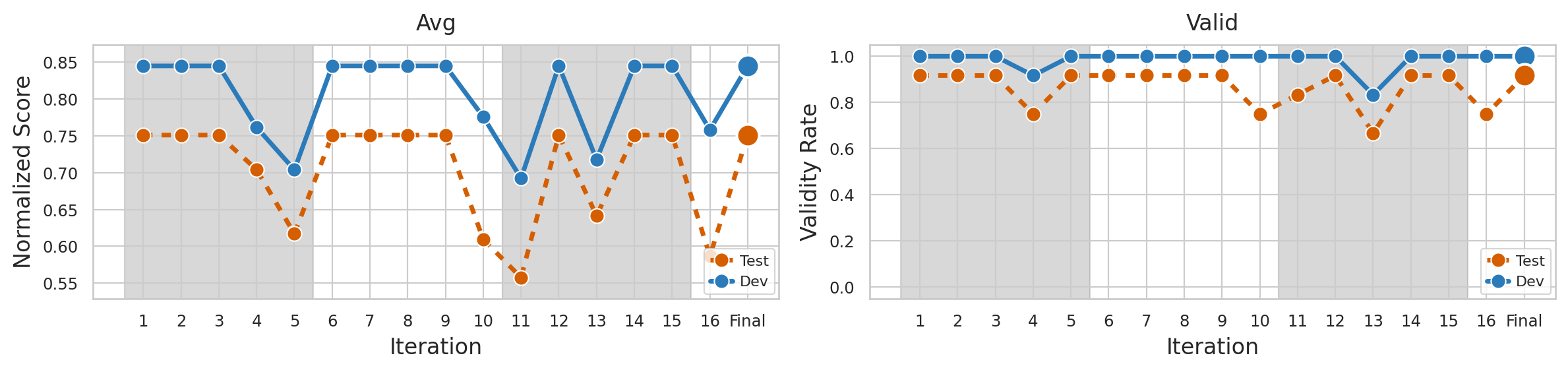}
  \caption{\textbf{Convergence on the development set} for the five CO problems not shown in \Cref{fig:convergence}: \emph{top to bottom,} Container Loading, Container Loading with Weight Restrictions, Crew Scheduling, Euclidean Steiner, and Resource-Constrained Shortest Path. Axes, shading, and the ``Final'' marker follow \Cref{fig:convergence}.}

  \label{fig:convergence_appendix}
  \vspace{-0.6em} 
\end{figure*} 

\clearpage

\end{document}